\documentclass{article} 
\usepackage{jfrExamplee}
\usepackage{graphicx}
\usepackage{apalike}
\usepackage{setspace}
\usepackage{booktabs}

\usepackage{nccmath}
\usepackage{subfigure}
\usepackage{cite}

\usepackage{gordon}
\usepackage{amssymb}

\newcommand{\gordonfirst}{\textcolor{black}}
\newcommand{\gordonpratap}{\textcolor{black}}
\newcommand{\revieweraddition}{\textcolor{black}}
\newcommand{\postrebuttal}{\textcolor{black}}

\newcommand{\wb}{\mathbf{w}} 
\newcommand{\Qb}{\mathbf{Q}} 

\begin{document}

\title{Radiation Search Operations using Scene Understanding with Autonomous UAV and UGV}

\author{
Gordon Christie\\
Virginia Tech\\
Blacksburg, VA \\
\And
Adam Shoemaker \\
Virginia Tech\\
Blacksburg, VA \\
\AND
Kevin Kochersberger \\
Virginia Tech\\
Blacksburg, VA \\
\And
Pratap Tokekar \\
Virginia Tech\\
Blacksburg, VA \\
\And
Lance McLean \\
Remote Sensing Laboratory\\
Joint Base Andrews, MD \\
\And
Alexander Leonessa \\
Virginia Tech\\
Blacksburg, VA \\
}





\maketitle

\begin{abstract}
Autonomously searching for hazardous radiation sources requires the ability of the aerial and ground systems to understand the scene they are scouting. In this paper, we present systems, algorithms, and experiments to perform radiation search using unmanned aerial vehicles (UAV) and unmanned ground vehicles (UGV) by employing semantic scene segmentation. \gordonfirst{The aerial data is used to identify radiological points of interest, generate an orthophoto along with a digital elevation model (DEM) of the scene, and perform semantic segmentation to assign a category (\eg road, grass) to each pixel in the orthophoto.}  We perform semantic segmentation by training a model on a dataset of images we collected and annotated, using the model to perform inference on images of the test area unseen to the model, and then refining the results with the DEM to better reason about category predictions at each pixel. We then use all of these outputs to plan a path for a UGV carrying a LiDAR to map the environment and avoid obstacles not present during the flight, and a radiation detector to collect more precise radiation measurements from the ground.
Results of the analysis for each scenario tested favorably. We also note that our approach is general and has the potential to work for a variety of different sensing tasks. 
\end{abstract}

\section{Introduction}
\label{sec:intro}

\begin{figure*}[ht!]
    \centering
    \includegraphics[width=\textwidth]{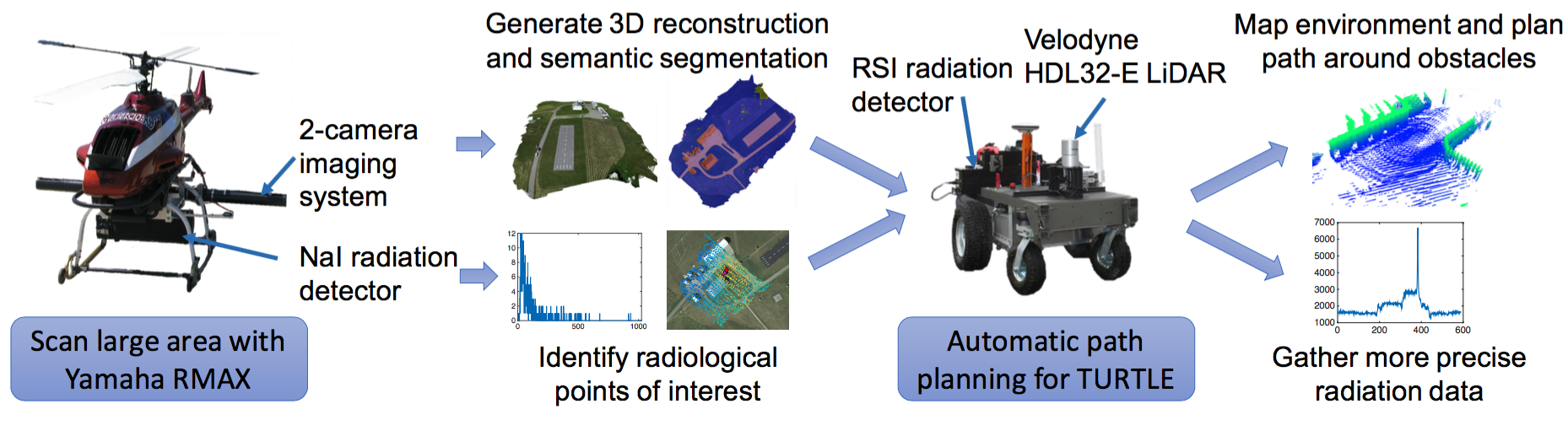}
    \caption{Overview of our approach to the autonomous search for radiation sources in an unknown environment.  The Yamaha RMAX is used to autonomously search a large area for radiation activity by collecting gamma radiation data.  By simultaneously collecting 2D color imagery, a 2D orthophoto and DEM can be generated for the area, which are then used to perform semantic segmentation. Using all image data outputs, a path can then be planned for a UGV, named the TURTLE, to collect more precise measurements around the point of interest. Since objects that were not present during the time of the flight may appear, LiDAR is used on-board the TURTLE to detect obstacles, which are then used to update a global map and find an alternate route.}
    \label{fig:teaser_fig}
\end{figure*}

Searches for illicit radiological or nuclear material, that would comprise a radiological dispersal device (RDD) or improvised nuclear device (IND), are becoming increasingly routine as commercial-off-the-shelf (COTS) radiation detection equipment has become more widely distributed among local, state and federal law enforcement and emergency response agencies. This increased capability comes at the cost of the time and personnel that must be allocated to the radiation/nuke search mission. Therefore, expediting the search process becomes paramount. The search process in general consists of the detection of anomalies, the localization of these anomalies and the identification of the sources of the anomalies (in this case radionuclides). Although the radiation data collected in this work can readily provide an unambiguous radionuclide identification, automated spectroscopic identification is not the subject of our research and has been well-studied elsewhere~\cite{jarman2008comparison,anderson2008discriminating}. Instead, we focus on detecting the anamolies using autonomous ground and aerial robots.

Autonomously searching for hazardous radiation sources provides a safer approach to what is possible via manned surveys. It can also be more efficient since a UAV is capable of autonomously scanning large areas to collect radiation data. Furthermore, existing maps for the area of interest may not be available or out-of-date. By taking images from a UAV it is possible to generate an updated 3D map of the area. Machine learning methods can be used to provide a semantic understanding of the scene that can be used to plan a path for a UGV to reach radiological points of interest. Once at the destination the UGV can then collect additional radiation data, transmit video to operators at a remote base station, and update the understanding of other unmanned systems simultaneously searching the area.

Performing this autonomous search in unknown environments is a challenging task.  In our approach to the problem we use a UAV and UGV to carry out the search missions. 
We use a Yamaha RMAX unmanned helicopter with an imaging system that takes 2D color images synchronized with GPS, and a \gordonfirst{Sodium Iodide (NaI)} radiation detector, designed and built by Sandia National Laboratories, to collect gamma radiation spectral data. The imagery collected from the RMAX is used to generate a 3D point cloud that can be processed into an orthophoto and DEM.  By performing semantic segmentation on this data to assign each pixel in the orthophoto with a semantic category, more intelligent reasoning can be used to plan a path for a UGV to visit the points of interest. The spectral data is analyzed to output these points of interest where sources are possibly located. 

While aerial scans are often capable of providing precise locations of radiation sources with high confidence of a source being present, this is not always the case. The scan lines in flight paths may not be dense enough for precise location estimates. Also, these location estimates may be at positions where no significant source of radiation exists.
We therefore use a UGV (the TURTLE), designed and built by \gordonfirst{The Center for} Dynamic Systems Modeling and Control (DySMAC) at Virginia Tech, to visit the points of interest on the ground. 
The TURTLE is equipped with a LiDAR, which generates a 3D point cloud to map the environment and detects and avoids obstacles while en route to the estimated source location to collect additional measurements from a radiation detector mounted on-board.
\postrebuttal{
We do not perform an active search of the area from the RMAX, which was done in \cite{kochersberger2014post}, since we consider the scenario where the UAV may be scanning a much larger area, and relies on one or more UGV to more closely inspect the scene.
}

\revieweraddition{
A lot of work has focused on the task of generating maps of an area from aerial imagery. There are several ways to to accomplish this task. One approach is to perform image stitching, where images are mosaicked together using feature matches to create a 2D image of the scene. Each pixel in the resulting map can then be georegistered if needed. While image stitching is fast and has many implementations available~\cite{BL07,Agrawal2015,microsoftice}, this does not provide 3D information, which is important to perform more accurate semantic segmentation and to plan better paths for UGV. Stereo vision provides another solution to this problem, where a calibrated two-camera imaging rig can be used to generate fast local 3D reconstructions from pairs of images. By reasoning about matching feature points in subsequent pairs of images, these local 3D reconstructions can be transformed into a global coordinate frame to create a full map of the area. There are several publicly available implementations for simultaneous localization and mapping (SLAM) from stereo vision~\cite{dellaert2012factor,Geiger2011IV,Badino:MVA11}, but these implementations typically require high frame rates to work well. The calibrated two-camera imaging rig that we use in our experiments has two low-cost point-and-shoot cameras that are not capable of high frame rates. While we are actively exploring ways of generating high-quality 3D reconstructions from the stereo pairs collected from our imaging rig, we use single-camera 3D reconstructions that are georegistered to obtain the aerial maps used in our experiments.
}

Although localizing radiation sources in an unknown environment is the primary motivation behind our work, the focus of this paper is providing autonomy to the aerial and ground systems collaborating to find them. \figref{fig:teaser_fig} shows an overview of our approach to the autonomous search for sources of radiation. 
\postrebuttal{The end goal that we have in mind is for our system to be able to autonomously identify the locations of potentially hazardous radiation sources with UAV and UGV. The UAV should be able to scan a large area to provide valuable context to a UGV that can efficiently search the area and confirm the presence of sources at the estimated locations. Although not presented in this work, future goals of this project include the ability of one or more UGV searching the area from the ground to scan the other areas in the scene to identify sources located at locations not identified by the UAV.}
References to work focused on finding sources from aerial data are provided in \secref{sec:related_work}. 
\postrebuttal{The main contribution of this work is a method to autonomously estimate and confirm the locations of radiation sources with UAV and UGV applying scene understanding in an unknown outdoor environment using aerial imagery with a supervised machine learning approach.  We also incorporate aerial semantic segmentation results into the A* path planning algorithm so that a UGV will prefer to follow roads over grass and stay clear of obstacles. We also demonstrate the ability to detect obstacles locally on the ground with LiDAR and then find a path around the obstacle using both local and global information.}

\vspace{-0.5cm}

\section{\gordonpratap{Related Work}}
\label{sec:related_work}

\paragraph{\revieweraddition{Unmanned Systems} Collaboration.}
\revieweraddition{
The collaboration between autonomous unmanned systems has been studied for a large number of applications. These unmanned systems include autonomous underwater vehicles (AUV), unmanned surface vehicles (USV), unmanned aerial vehicles (UAV), and unmanned ground vehicles (UGV). Some examples of the applications of these unmanned systems are search and rescue operations, post-disaster surveying, target localization and tracking, and precision agriculture monitoring. Previous works have focused on the collaboration between multiple UAV~\cite{yu2013formation, pestana2014vision, kushleyev2013towards, dong2015time}, multiple UGV~\cite{bruggemann2012outdoor, lim2009formation, anisi2008cooperative, hussain2004evolution, deusdado2016aerial}, the collaboration between UAV and UGV~\cite{duan2014multiple, kim2014multi, tokekar2013sensor, cheung2008uav, phan2008cooperative}, and much more.
Garzon \etal present a solution for multiple UGV to perform signal searching tasks in large outdoor scenarios~\cite{garzon2015multirobot}. They propose different path planning strategies for coverage, which depend on the size and shape of the field.
}
\postrebuttal{
Careful consideration of the aerial search patterns is important for extending our work. Currently, we use equally spaced scan lines, where the distance between the scan lines is based on the expected overlap to obtain a high quality 3D reconstruction of the environment.
}

For the topic of UAV-UGV collaboration, Tokekar \etal studied the problem of coordinating UAV and UGV for precision agriculture~\cite{tokekar2013sensor}, where they found energy efficient ways to visit areas with misclassified nitrogen levels. 
UAV and UGV have also been used in a collaborative manner to perform target localization \cite{tanner2007switched,grocholsky2006cooperative}.  
In~\cite{mueggler2014aerial} a mock-up disaster scenario was setup, where a UAV maps the area and then computes the fastest mission for a UGV to reach the destination and deliver a first-aid kit. Cooperative environment mapping~\cite{michael2012collaborative} and surveillance~\cite{saska2012cooperative} have also been studied. In our experiments we are interested in simply reaching a destination and returning to the start position, but note that our semantic segmentation results could be used to find better paths to take for such tasks in outdoor environments.
While our experiments are farily specific, and therefore difficult to compare to existing approaches, Schneider \etal discuss how EURATHLON and ELROB have provided a way of standardizing and benchmarking the evaluation of methods in outdoor robotics through competition~\cite{schneider2015elrob}. 
\postrebuttal{
Teams at these competitions build impressive systems that are capable of executing missions in real-time for important tasks such as search and rescue. 
}
\postrebuttal{
Others have used overhead imagery to improve UGV path planning capabilities. In \cite{sofman2006improving}, a self-supervised online learning algorithm is used on a UGV to learn a model that integrates information about the current terrain and overhead imagery that is then used to predict traversal costs at other regions in the overhead map. These predicted traveral costs were then used to perform path planning.
}
\postrebuttal{
While many of these works demonstrate successful collaboration between UAV and UGV, we try to focus more on using semantic segmentation for scene understanding in a real-world search task by training on a dataset of imagery annotated with semantic categories. As more images are captured and annotated by low-flying aircraft, we believe it will be important to integrate existing models with online learning algorithms, such as the one presented in \cite{sofman2006improving}. These models will be able to provide valuable context to a UGV during tasks such as radiation search, as existing maps (\eg satellite) may be too old to capture important information about the scene.
}

\paragraph{Scene Understanding.}
Perception for autonomous robotic systems has seen tremendous progress in many applications. A variety of possible sensing methods (RGB-depth sensors, visual cameras, acoustic sensors, LiDARs, \emph{etc.}) have allowed these systems to perceive the world and make intelligent decisions. 
Semantic segmentation has been the focus of many works, with state-of-the-art models~\cite{chen14semantic,lin15efficient} capable of achieving high accuracy for many different tasks, and large datasets with semantic annotations available for training and evaluation~\cite{silberman2012indoor,pascalvoc2012}.  
However, to the best of our knowledge, no publicly available dataset of semantically annotated images from low-flying UAV currently exists.  In this work we create our own dataset to train a model to perform semantic segmentation with 2D color images and ground truth annotations, then evaluate on unseen image tiles of the orthophoto for the area in which we are searching for hazardous radiation sources. 
\postrebuttal{
We believe that this approach will extend to similar scenes with the same set of categories present in the scenes. As aerial datasets grow, and as more images are annotated, we will be able to take advantage of segmentation algorithms such as DeepLab-CRF~\cite{chen14semantic} to more accurately segment a wide variety of scenes. 
}

Similar to this work, Montoya-Zegarra \etal explored road mapping~\cite{montoya2014mind} and the semantic segmentation of aerial images with higher-order cliques~\cite{montoya2015semantic}. 
\postrebuttal{
In our work we are more focused on segmenting images captured from low-flying UAV, the outputs of which need to be used for planning UGV missions to search for sources of radiation.
}
Radford studied the problem of real-time roadway classification from aerial imagery for UGV path planning~\cite{Radford2014MS}, where k-means clustering and image mosaicking were used.  This approach, however, relies on an initialization step where the algorithm is first shown which cluster is a road.  Our approach uses supervised learning to perform semantic segmentation of aerial imagery for several categories, which tends to scale well and requires no human supervision at test time. 
Supervised classification of LiDAR point clouds has also been studied~\cite{Xiong_2011_6787,niemeyer2014contextual}. 
\revieweraddition{
Joint semantic segmentation of 2D and 3D data simultaneously has been the focus of several other works. Floros \etal presented an approach to perform semantic segmentation of 2D images and 3D point clouds generated from stereo pairs with a joint model that incorporated temporal consistency between subsequent frames~\cite{floros2012joint}. Munoz \etal developed an approach to jointly perform semantic segmentation of 2D images and 3D LiDAR point clouds by integrating information between overlapping parts of the scene~\cite{munoz-eccv-12}.
In a work by Sturgess \etal, structure from motion features were incorporated into the semantic segmentation of road scenes~\cite{sturgess2009combining}.
}
\postrebuttal{
While all of these works are relevant to our paper, we do not focus too much on a framework to perform joint 2D+3D semantic segmentation of the aerial data, which would require more training data than we have available to us. We also do not focus on performing better semantic segmentation of the scene from the UGV data. However, we do note that these ideas are interesting directions for future work. 
}
\postrebuttal{In our work,} the LiDAR on the TURTLE is used to detect obstacles on its current path by analyzing elevation gradients, which we found sufficient for our task. 
\postrebuttal{
For the semantic segmentation of the orthophoto generated from the imagery captured by the RMAX, we have a two-stage approach where we analyze the DEM separately to make better category predictions at each pixel. Ideally, we would implement a joint framework, such as the ones presented in \cite{floros2012joint,munoz-eccv-12}, but we do not have enough aerial data for this.
}

\revieweraddition{
Yingze \etal presented an approach to generate image-based 3D reconstructions while recovering the locations, poses, and categories of objects in a scene~\cite{bao2011semantic}. 
In a work by Kundu \etal, an approach was presented for joint inference of 3D scene structure and semantic segmentation of urban street scene imagery~\cite{kundu2014joint}. 
\postrebuttal{
We would require more data to be able to use these types of approaches with the imagery collected by the RMAX, but note that this is also an interesting direction for future work. In this paper, we find that the segmentation output by the model we train is very acceptable for our given application.
}
Incorporating semantic maps into path planning for mobile robots has also been studied.  Hatao \etal proposed a semantic map making system based on road structures, where trajectories of moving objects, landmarks, building entry points, and traffic signs are added to the map~\cite{hatao2014construction}. They combine laser range finders with an omnidirectional camera for perception on the robot.
}
\postrebuttal{
In this work we develop road structures as part of the segmentation process for our orthophoto, which are used by the UGV to plan global paths to the destinations with possible sources of radiation.
}

Others have also studied optimal camera positions for UAV collecting imagery to be used for image-based 3D reconstructions~\cite{rs8010026}. While this is ideal for generating a better orthophoto and DEM, navigating to 3D positions not on scan lines with the same altitude increases the amount of time to complete the mission, and makes analyzing the radiation spectral data more difficult. We therefore use scan lines when planning the missions for the RMAX.

\paragraph{Radiation Sensing.}
There has also been research on using UAV and UGV for radiation mapping missions.  Kochersberger \etal studied mapping radiation levels in an unknown environment using a UAV to collect radiation data from the air and deploy a tethered UGV to collect samples from the ground~\cite{kochersberger2014post}.  
\postrebuttal{While similar to this work, their work focused on active radiation search strategies with no focus on the planning for the tethered UGV to reach the destination. In this paper, we present a full system that performs an analysis of the radiation data after the UAV lands, plans a path for a UGV to visit points of interest collecting additional radiation data while avoiding obstacles on the way to the destination. We also use vision-based scene understanding to complete the missions, which allows for low-cost cameras to be used.}
Vetter \etal use an RMAX to map radiation and propose a ``Nuclear Street View''~\cite{vetter2015advanced}.
Schneider \etal discuss possible scenarios for collecting radiation measurements with unmanned systems~\revieweraddition{
\cite{schneider2015possible}, where one type of scenario is the prevention of incidents involving radiation and the other post-incident analysis. Our experiments focus on the prevention scenario\postrebuttal{, where we perform a scan of a large area assuming all parts of the scene are equally important. We simultaneously develop a semantic understanding of the scene which helps us to plan a mission for a UGV to then visit the areas of activity and potentially prevent a distaster from occuring}.
}
Benedetto \etal developed an approach to identifying regions of interest in radiation data by means 
of clustering that is driven by diffusion operators as applied to a data graph representation of the collection of radiation spectra~\cite{benedetto2014spie}. While more 
advanced reasoning could easily be incorporated into our search, such methods are not necessary to demonstrate
the successful automation of the process of finding and localizing radiation anomalies. We find that the use of a simple approach based on the local maxima in the overall intensity (calculated as the sum of of the counts in all spectral channels for each measurement) to indicate potential source locations works well in our experiments. 
\postrebuttal{
The approach by \cite{benedetto2014spie} provided equal performance to the use of max counts in our experiments, but we note that for more complicated experimental setups, using such an approach will become necessary.
}
We instead focus more on augmenting semantic information into the search process.
In another work by Schneider \etal \cite{schneider2012unmanned}, a prototype of an unmanned multi-robot reconnaissance system to detect chemical, biological, radiological, nuclear, and explosive (CBRNE) threats was presented, where the environment is not known a priori. Chemical and biological samples are obtained from the environment, and path planning is also performed so that trajectories can be generated to avoid obstacles. 
\postrebuttal{
In this work, we also avoid obstacles identified from the UGV, but incorporate both global and local information about the surrounding area to plan a path around them. The orthophoto segmentation that we generate also provides valuable context to the UGV mission to create an efficient initial path for the UGV to visit the regions of interest.
}

\revieweraddition{
Strategies for radiation search have also been explored. Cortez \etal propose two different motion planning strategies for building a radiation map~\cite{cortez2008smart}. One involves searching areas with higher uncertainty levels, and another involves visiting all cells in a grid where the amount of time spent at each cell depends on the uncertainty. Minamoto \etal estimate the intensities of radiation sources on the ground surface in 3D using a dosimeter~\cite{minamoto2014estimation}. By moving the dosimeter around in 3D, they perform a MAP estimation of the source intensities by using characteristics of attentuation. In the work by Towler \etal, present a grid-based robust Bayesian estimator to localize a single radiation source, and a contour analysis technique to localize an arbitrary number of radioactive sources~\cite{towler2012radiation}. All of these experiments were completed using simulated data. Brewer proposed a control strategy for a Yamaha RMAX unmanned helicopter to search for radiation sources using particle swarm particle filtering~\cite{brewer2009autonomous}.
}
\postrebuttal{
We believe that these approaches will become important in future work, where more complicated source configurations are used in more complicated environments. Instead of planning a path to visit one potential source location, we will be able to optimize a path to visit multiple locations, while performing active search along the way.
}

\section{\gordonpratap{Overview of the Method}}
\label{sec:overview}

Our method to autonomously search for hazardous radiation sources in an unknown outdoor environment uses a UAV (Yamaha RMAX) and UGV (TURTLE) to collaboratively understand the scene. We perform two separate missions
in two separate adjacent areas of Kentland Farm, Blacksburg, VA, where in the first mission we set up a single radiation source location, and in the second mission we set up two source locations. The RMAX missions are planned by using sets of scan lines.  The goal of each mission is to find and confirm the existence and locations of anomalous radiation sources. The TURTLE missions are planned automatically using outputs from the RMAX missions, where the start positions were arbitrarily set to the edge of the map on one of the roads entering the scene. 

The RMAX carries an imaging system to take 2D color images and a radiation detector to collect gamma radiation data. We use the images from both missions to create an orthophoto and DEM\footnote{The orthophoto and DEM are the same size, and when overlaid on one another represent the same part of the scene at each pixel.} for the combined flight areas to plan paths for the TURTLE, but treat the radiation data separately for each mission. The orthophoto and DEM are used to perform semantic segmentation.  We train a segmentation model on a dataset of images that we annotated with different categories (road, grass, building, vehicle, vegetation, and shadow) at each pixel, where the images in this dataset were taken from low-flying UAV in a variety of environments. This semantic segmentation is used for planning paths for the TURTLE.
The NaI radiation detector on-board the RMAX is used to estimate locations of potential sources. 
The radiation spectral data is output in the form of 1024-d vectors, where the sum of these vectors is called the counts.  Stronger sources can typically be found by looking only at the counts, but for weak sources located near stronger sources, more advanced reasoning is typically required. In our experiments, we use the simpler approach of using counts. 
\postrebuttal{
We also note that the max counts value that is found is a global maximum, meaning that only one source per scan be found with this approach.
}

\revieweraddition{
The spectral data that is output as 1024-dimensional vectors are synchronized with GPS to provide geospatial information about each detector reading. The source locations in each mission are estimated from the aerial data by the GPS position associated with the maximum counts (sum of the 1024-d vectors).
}
To confirm that the radiological points of interest from the aerial data actually contain a potentially hazardous source of radiation, we use the estimated source locations in the discrete set of aerial measurements as destinations for the TURTLE to visit in each mission. For the TURTLE to visit these points, we use the orthophoto, DEM, and segmentation to intelligently plan a path that prefers roads and keeps a safe distance from obstacles. An RSI 701 radiation detector\footnote{The RSI 701 is a different radiation detector to the NaI radiation detector mounted on the RMAX.} is mounted to the TURTLE to collect additional measurements around the estimated location. Since the scene may change between the end of the flight and the beginning of the ground operation, the TURTLE is equipped with LiDAR to identify obstacles and send coordinates bounding the obstacle to the global path planner to find an alternate route to the destination. LiDAR scans are also used to build a global map of the scene. \figref{fig:teaser_fig} provides an overview of the aerial and ground operations to perform the search.

We provide details for each step of the method in the following sections, which are organized as follows: \secref{sec:unmanned_systems} provides details of the Yamaha RMAX, the TURTLE, and their hardware. Details of the the image-based scene understanding, including the 3D reconstruction and semantic segmentation of the aerial imagery, are presented in \secref{sec:image_understanding}. 
\secref{sec:path_planning} discusses the path planning for the TURTLE to visit points of interest and how the semantic segmentation is incorporated. 
In \secref{sec:experiments} we present our experiments for both the RMAX and TURTLE missions. Finally, our thoughts on the experiments and potential future work are presented in \secref{sec:conclusions}.

\section{Unmanned Systems}
\label{sec:unmanned_systems}

In this section we detail the unmanned systems used to complete all of the experiments presented in this paper.

\subsection{Unmanned Aerial Vehicle -- Yamaha RMAX}
\label{sec:rmax_description}

\begin{figure}[h!]
    \centering
    \includegraphics[width=0.4\columnwidth]{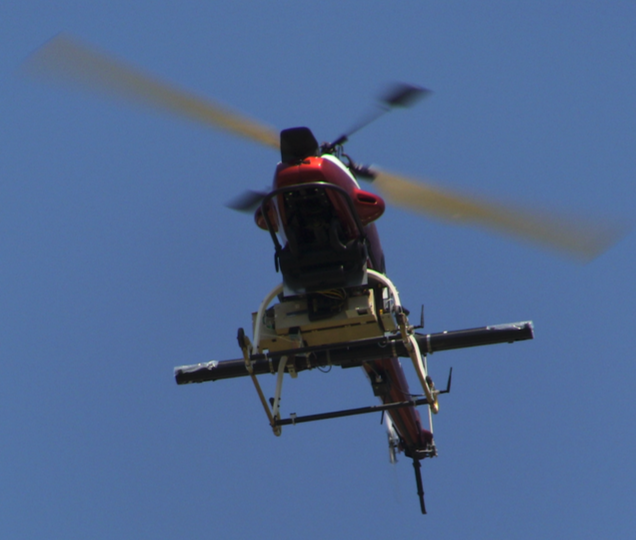}
    \caption{The NaI radiation detector and imaging system mounted to the RMAX during one of the missions.}
    \label{fig:rmax_hardware}
\end{figure}

The UAV used is a 2005 Yamaha RMAX (model: L17-2), an aircraft originally developed for crop dusting in Japan. The wePilot autopilot system is used to interface with the flight control system and ground control allowing for autonomous operation. The RMAX has a 94kg gross weight, a max payload capacity of 28kg, and flight endurance time of approximately 45 minutes. The RMAX is shown in \figref{fig:rmax_hardware} during one of the missions carrying the radiation detector and imaging system.

\begin{figure*}[ht!]
    \centering
    \subfigure[Counts histogram for background with no radiation sources near detector ($\mu = 436.1$, $\sigma = 20.8$, $N = 600$).]{
		\label{fig:counts_hist_bg}
		\includegraphics[width=0.48\columnwidth]{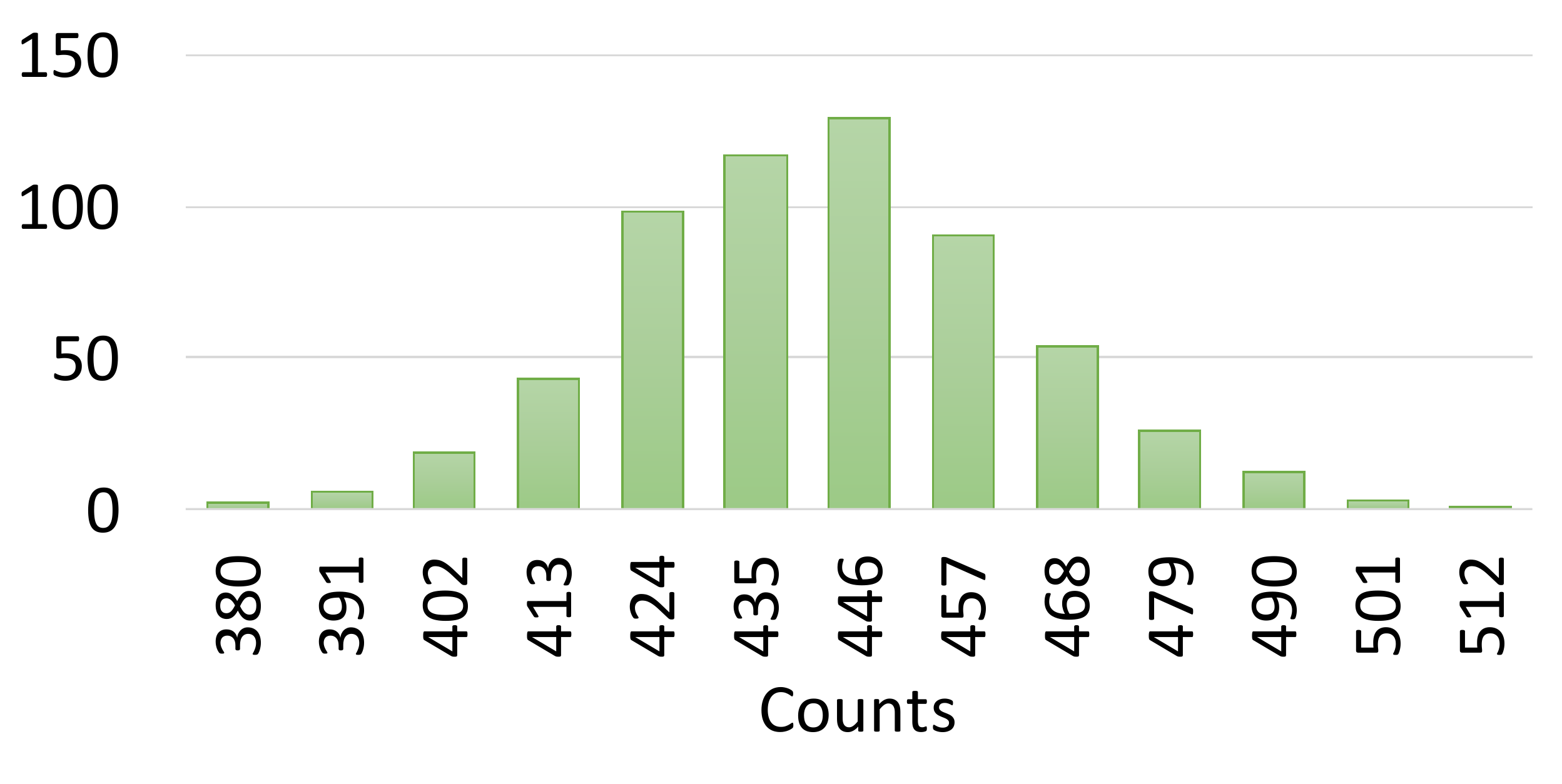}
	}
	\subfigure[Counts histogram with radiation source next to detector ($\mu = 739.7$, $\sigma = 28.5$, $N = 600$).]{
		\label{fig:counts_hist_source}
		\includegraphics[width=0.48\columnwidth]{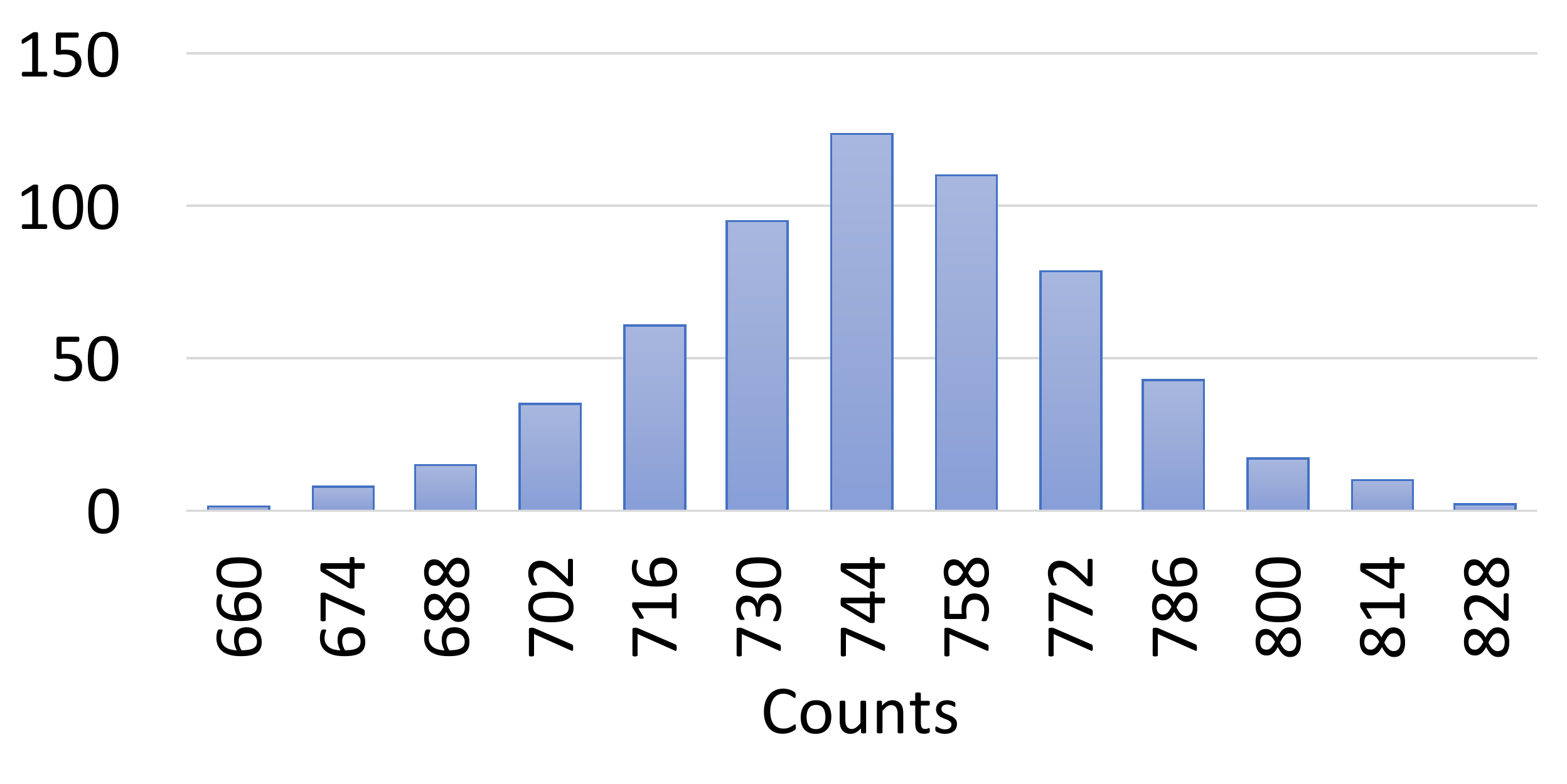}
	}
    \caption{Histograms (not normalized) of the counts from the NaI radiation detector mounted to the RMAX over a period of 10 minutes for (a) background measurements and (b) with a radiation source ($^{137}$Cs) present for calibration.}
    \label{fig:histograms_nai_detector}
\end{figure*}

\paragraph{Radiation Detector and Imaging Hardware.}
The radiation detector used to collect radiation spectral data is an NaI scintillation-type detector with a 9in length and 3in diameter. In order to understand the measurements of the detector during the missions, we first take background measurements and measurements with a $^{137}$Cs radiation source next to the detector for 10 minutes each. Histograms for each case are shown in \figref{fig:histograms_nai_detector}.

The imaging system mounted on the RMAX is a two-camera stereo boom designed and built by the Unmanned Systems Lab at Virginia Tech\footnote{Although we use a two-camera system, the orthophoto and DEM used in our experiments were generated from images from only one of the cameras.}.
Two off-the-shelf Canon PowerShot A-810 cameras were placed inside a carbon fiber tube resulting in a 1.38m baseline.  External power is provided to each camera, eliminating the need to remove the cameras for battery replacement, which would require aligning the cameras and performing a stereo calibration after each replacement.  In addition, SD cards are attached with extension cables that allow for quick mounting and dismounting. In order to synchronize the triggering of the cameras, we use a microcontroller that sends pulses over the USB power line and the Stereo Data Maker firmware~\cite{stereodatamaker}.

\subsection{Unmanned Ground Vehicle -- TURTLE}
\label{sec:turtle_description}

DySMAC designed and built four identical UGV referred to as the Terrestrial Unmanned Robots for Teamed Learning and Exploration (TURTLEs), one of which is used in our experiments. The control strategy to navigate the waypoints from the global planner was developed in~\cite{shoemaker2015bioinspired}.
The base design of the TURTLE includes a differential drive system, powered by two brushless motors located in the rear.  Each motor can run continually at speeds up to 10 mph (4.5 m/s) and produce torque up to 322 in-lbs.  These specifications, along with four wheel independent suspension, allow for traversal over a wide variety of terrains in both urban and rural environments. Moreover, the vehicle has been tested with payloads up to 100 lbs, a feature which allows for the deployment of the radiation detector and Velodyne HDL32-E LiDAR mounted on-board.

\begin{figure*}[ht!]
    \centering
    \subfigure[]{
		\label{fig:turtle_mid_mis}
		\includegraphics[height=2.2in]{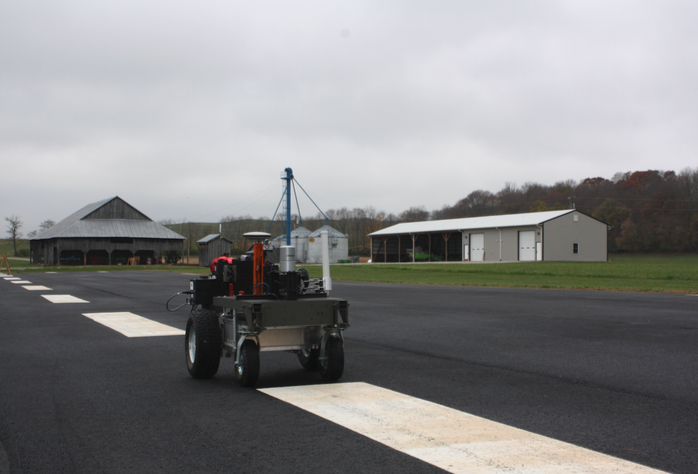}
	}
	\subfigure[]{
		\label{fig:rsi_detector}
		\includegraphics[height=2.2in]{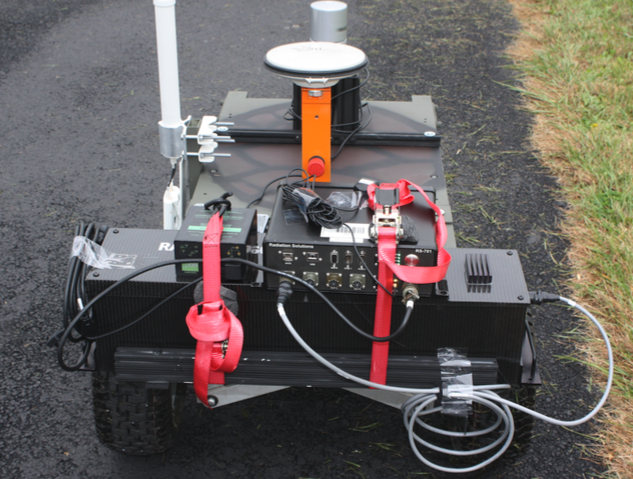}
	}
    \caption{(a) The TURTLE at Kentland Farm, Blacksburg, VA where all experiments took place. (b) RSI 701 radiation detector mounted on the back of the TURTLE.}
    \label{fig:turtle_image}
\end{figure*}

\gordonfirst{The TURTLE contains an on-board computer, 5 GHz radio, and GPS/INS system. 
The TURTLE's computer has an i-7 Intel processor, 80 GB SSD, and 8 GB of RAM. This allows for full
vehicle control and sensor collection along with building a global DEM in real-time.
The radio establishes high bandwidth full inter-vehicle communication, which can broadcast over several miles, facilitating wide scale implementation. This network can easily be augmented to include personnel communication as well. Using this network, processing can easily be distributed on a need-be basis. Moreover, the network allows for clear position knowledge from every other unit, strengthening the estimate.
The built in NovAtel SPAN-CPT GPS/INS system is rated up to a position accuracy of 1m.} 
\postrebuttal{This, without SLAM, was enough to achieve acceptable global LiDAR maps.}
The TURTLE used is shown in \figref{fig:turtle_image} at the Kentland Farm test area. 
\revieweraddition{
The computer runs Windows 7, where LabVIEW is used for all control of the robot and processing of the LiDAR data.
}

\revieweraddition{
There are several reasons to send in a UGV for further inspection. One reason is that the UAV may be performing a scan of a larger area, where scan lines are not that dense. Having a UGV inspect the scene can provide a more precise estimate of the location, as it can get closer to the source. The UGV may also be able to visit areas that are difficult for a UAV to reach. It is also possible to perform long-dwell measurements and to carry larger detectors with higher sensitivity on the TURTLE than the RMAX, which allows for the collection of statistically better data and better localization. 
}

\begin{figure*}[ht!]
    \centering
    \subfigure[Counts histogram for background with no radiation sources near detector ($\mu = 1469.4$, $\sigma = 42.7$, $N = 600$).]{
		\label{fig:counts_hist_bg}
		\includegraphics[width=0.48\columnwidth]{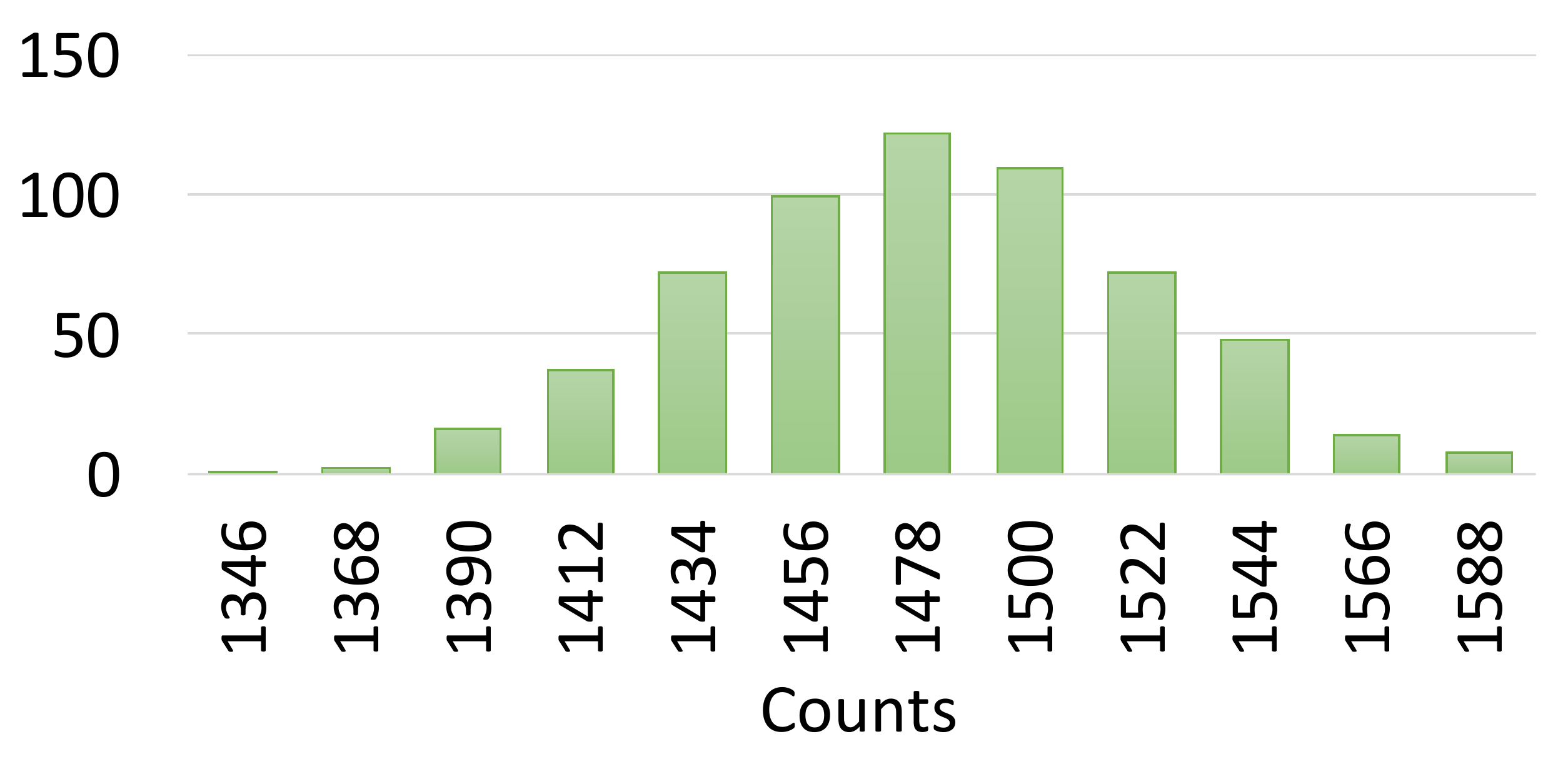}
	}
	\subfigure[Counts histogram with radiation source next to detector ($\mu = 1956.5$, $\sigma = 43.4$, $N = 600$).]{
		\label{fig:counts_hist_source}
		\includegraphics[width=0.48\columnwidth]{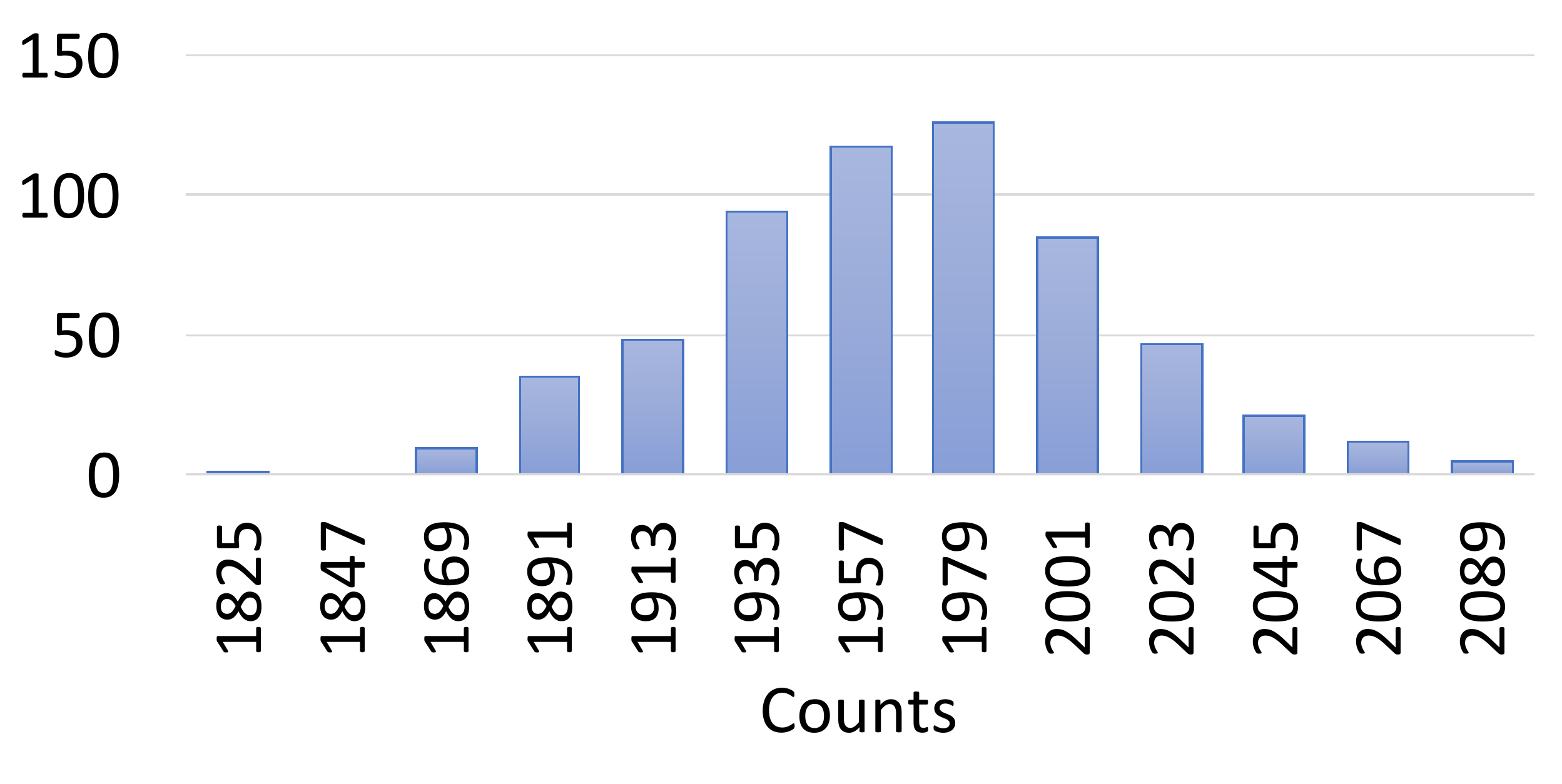}
	}
    \caption{Histograms (not normalized) of the counts from the RSI 701 radiation detector mounted to the \revieweraddition{TURTLE}  over a period of 10 minutes for (a) background measurements and (b) with a $^{137}$Cs radiation source present for calibration.}
    \label{fig:histograms_rsi_detector}
\end{figure*}

\paragraph{Radiation Detector}
The radiation detector is a 2x4x16 inch NaI(Tl) system, manufactured by Radiation Solutions, Inc (RSI) that is shown mounted to the back of the TURTLE in \figref{fig:rsi_detector}. 
Similar to the Sandia system the RSI detector records second-by-second gamma-ray spectra into GPS-tagged, 1024-channel histograms that span an energy range of 0-3000 keV. 
We performed a calibration of the detector by taking background measurements and measurements with a $^{137}$Cs radiation source placed next to the detector for 10 minutes each. A histogram of the counts for each case is shown in \figref{fig:histograms_rsi_detector}. 

%

\section{Image-based Scene Understanding}
\label{sec:image_understanding}

\gordonpratap{Having a 3D reconstruction of the scene is necessary to be able to reliably perform segmentation and plan a path for a UGV. Using only 2D information from the images to plan a path can fail when the segmentation algorithm used confuses the traversable and non-traversable categories it is segmenting.} For our aerial operations we chose to use 2D color cameras for perception, as they provide a reliable and low-cost solution compared to LiDAR \gordonpratap{for generating 3D reconstructions}.  We were also able to complete all experiments using off-the shelf Canon PowerShot cameras, \gordonfirst{for which there is no noticeable loss of accuracy} in the 3D reconstructions when compared to expensive machine vision cameras previously tested over Kentland Farms in the same flight area.  The 2D color images were also proven to be useful for performing semantic segmentation, especially when distinguishing between categories with similar elevation patterns, such as grass and roads.

\subsection{3D Reconstruction}

We considered two different methods of image-based 3D reconstructions in this work. The first method is stereo vision, where 3D positions of pixels matched between the left and right images are calculated using a calibration of the imaging system.  Advantages of this approach include fast computation and the ability to track dynamic parts of the scene in 3D.  The second approach is using structure from motion, where a 3D point cloud is generated by reasoning about pixels matched between two or more images.  This results in a more accurate 3D reconstruction than with stereo vision because the depth resolution is increased by viewing most of the points from more than two camera positions.  However, structure from motion is usually much slower than stereo vision, as this now involves optimizing for the 3D position of each point using the pixel positions from all images it is visible within, and also optimizing for the camera positions.  Dynamic parts of the scene are also difficult to model with this approach, which typically results in their absence from the final 3D reconstruction.  \postrebuttal{An advantage of structure from motion over stereo vision is that it tends to create a more accurate orthophotos and DEM}, which is useful for applications such as path planning, which we explore in this paper.  When attempting to stitch local stereo reconstructions together, we found the results \postrebuttal{much more noisy than what Agisoft output, which were significantly cleaner and more accurate. This allows for better obstacle detection, which is used to segment the scene. A natural drawback of structure from motion, however, is the inherent scale ambiguity associated with a monocular setup in the absence of GPS. In GPS-denied areas it is better to use stereo vision, as it is capable of providing 3D reconstructions with known scale.}

For structure from motion we tested two different implementations.  The first implementation tested was VisualSFM~\cite{visualsfm2013,wucvpr2011}, which we combined with a multi-view stereo implementation, PMVS~\cite{furukawa2010accurate} to generate a dense 3D reconstruction after initializing itself with the sparse reconstruction output by VisualSFM.  We also tested the professional edition of Agisoft~\cite{agisoft2013}.  Of the two, Agisoft provided superior results out-of-the-box, and has the additional capability of generating orthophotos and DEMs, which are more convenient inputs to path planning algorithms.  

\begin{figure}[h!]
    \centering
    \includegraphics[width=0.7\columnwidth]{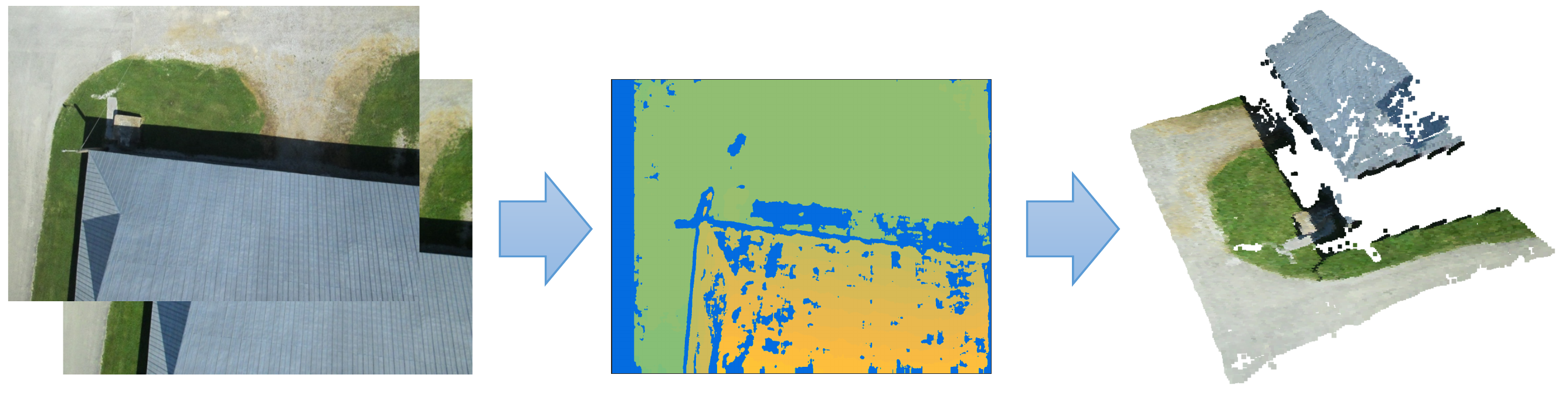}
    \caption{Example stereo vision 3D reconstruction, with example image pair (left), disparity map calculated using example images (middle), and the 3D reconstruction generated from the disparity map and calibration file (right).}
    \label{fig:disparity_flow}
\end{figure}

In this paper, we collected stereo images, but images from only one camera were used to generate 3D reconstructions from SfM. In future work, stereo vision could potentially be used for real-time reasoning about possible radiation source locations.  To generate a stereo reconstruction, we first undistort and rectify the images of the left and right cameras using a calibration file generated using the MATLAB Calibration Toolbox~\cite{bouguet2004camera}. To calculate the disparity map for the image pair, we then use the semi-global block matching algorithm provided by OpenCV~\cite{bradski2008learning}. \figref{fig:disparity_flow} shows the process of generating a 3D point cloud from a pair of stereo images taken from the Canon A-810 cameras during one of the RMAX missions.  The disparity map is first calculated, after which the calibration data is used to generate 3D points for each pixel with a valid disparity value as

\begin{equation}
w \begin{bmatrix}
    X_C \\
    Y_C \\
    Z_C \\
    1
\end{bmatrix}
=
\Qb \begin{bmatrix}
    x \\
    y \\
    d \\
    1
\end{bmatrix}
=
\begin{bmatrix}
    1 & 0 & 0 & -C_x \\
    0 & 1 & 0 & -C_y \\
    0 & 0 & 0 & f \\
    0 & 0 & -\frac{1}{B} & \frac{(C_x-C'_x)}{B}
\end{bmatrix}
\begin{bmatrix}
    x \\
    y \\
    d \\
    1
\end{bmatrix}
\end{equation}

where $(X_C,Y_C,Z_C)$ is the 3D point with units in meters, ($x$,$y$) is the pixel position, and $d$ is the disparity value. The back projection ($\Qb$) matrix is generated using the calibration and is used to transform the local pixel coordinates and disparity value to a 3D point. In this matrix, $(C_x,C_y)$ is the camera center, $f$ is the focal length, and $B$ is the baseline distance between the cameras, where all parameters are from the left camera except for $C'_x$~\cite{bradski2008learning}. 
It was observed that cleaner reconstructions were possible by convolving the depth image with a median filter. We found the results from local stereo reconstructions to be impressive given that these are off-the-shelf cameras with retractable lenses taking images while mounted to the RMAX, which causes a lot of vibration.

\postrebuttal{
We chose to use Agisoft in this work, which typically takes multiple days of computation to reconstruct the an area similar in size of Kentland Farm. The orthophoto and DEM used in our work help to demonstrate the capabilities of our system. However, our approach does not rely on Agisoft, and our work does not intend to improve upon existing reconstruction methods. If more expensive machine vision cameras are used than the point-and-shoot cameras used in our work, then a real-time 3D reconstruction of the area can be generated using existing code \cite{dellaert2012factor,engel2014lsd,mur2015orb}. If real-time 3D reconstructions are provided, then a real-time response by our system would be possible.
}

\subsection{Aerial Dataset and Semantic Segmentation}

\gordonfirst{Some researchers performing segmentation of aerial imagery use unsupervised approaches, where no training data is used to help make predictions~\cite{Radford2014MS,lin2012road}. These often fail due to assumptions of good initializations that subsequent segmentations rely on (\eg operator labeling the road in the first image of a road-tracking UAV) and arbitrary hand-crafted parameters (\eg RGB thresholds for classification) that do not extend to other scenes.}  For this reason, we decided to take a supervised approach to the problem, where we train a segmentation model to predict one of several semantic categories for each pixel in an image.  This approach requires no initializations of any kind, and is also scalable, since no algorithm changes are required when testing on a different type of scene.  In the case the approach does fail, then it is likely that there is a simple need for more training data.

\begin{figure*}[ht!]
    \centering
    \includegraphics[height=4.55cm]{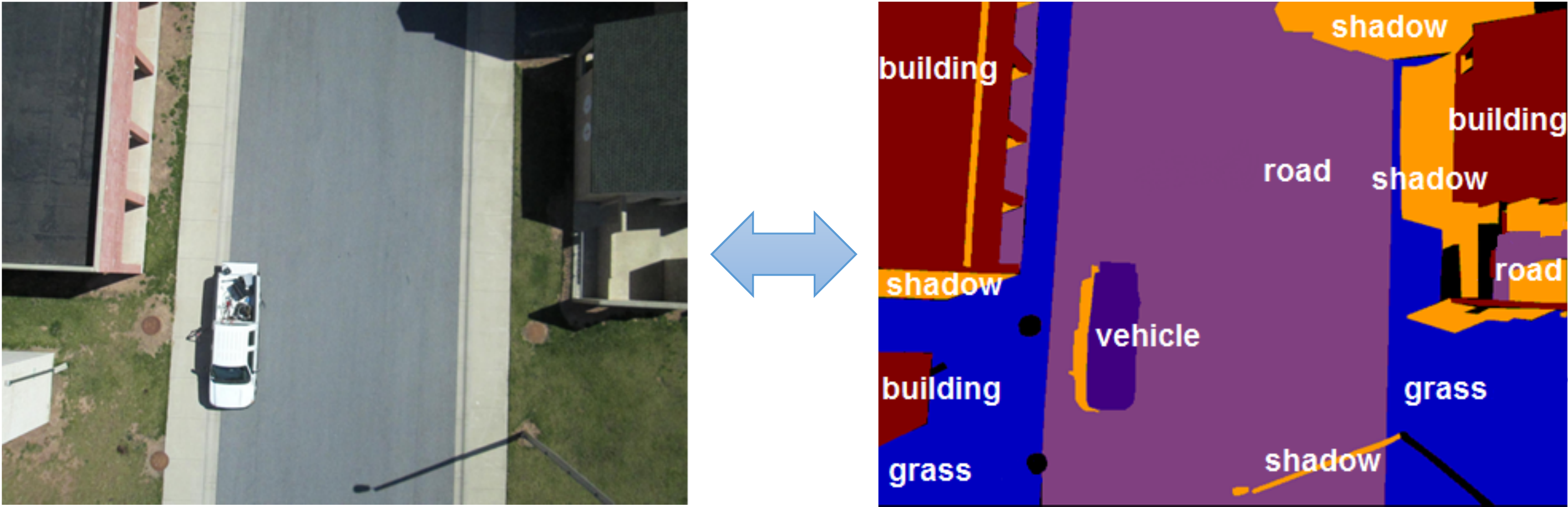}
    \hspace{.11in}
    \includegraphics[height=4.55cm]{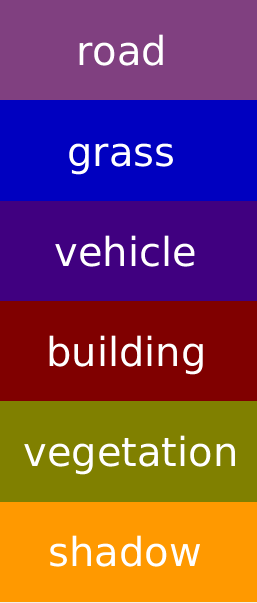}
    \caption{We annotate 2D RGB images taken from low-flying UAV with 6 different semantic categories that we use to train a model to predict the categories for pixels of unseen test images.}
    \label{fig:aerial_dataset_overview}
\end{figure*}

For the work presented in this paper, we annotated a collection of images taken from low-flying UAV in a variety of environments with several semantic categories to be able to train the segmentation model that predicts these categories on the unseen test images of Kentland Farms. The images were annotated using LabelMe~\cite{labelme2007}.  \figref{fig:aerial_dataset_overview} shows an example annotation from our dataset and the legend for the colors for each category. The full dataset consists of 230 annotated images, where 54 come from tiles of the orthophoto for the Kentland Farms flight, 119 come from an RMAX flight conducted by the Unmanned Systems Lab at Virginia Tech in Fort Indiantown Gap, PA, and 57 come from a variety of flights taken from low-flying UAV.  However, for training, we only use a subset of the 119 Fort Indiantown Gap images to prevent the model from overfitting to this scene.  We use 15 images from this part of the dataset, resulting in a total of 72 training images when testing on the orthophoto of the Kentland Farms imagery. 
\revieweraddition{
Ideally we would collect a very large dataset with more semantic categories so that a deep semantic segmentation model, such as DeepLab-CRF~\cite{chen14semantic}, could be used. However, collecting such a dataset is difficult, since a diverse set of images from low-flying UAV are not easy to find, and the annotation procedure is exensive. 
}

\begin{figure*}[ht!]
    \centering
    \includegraphics[width=1\textwidth]{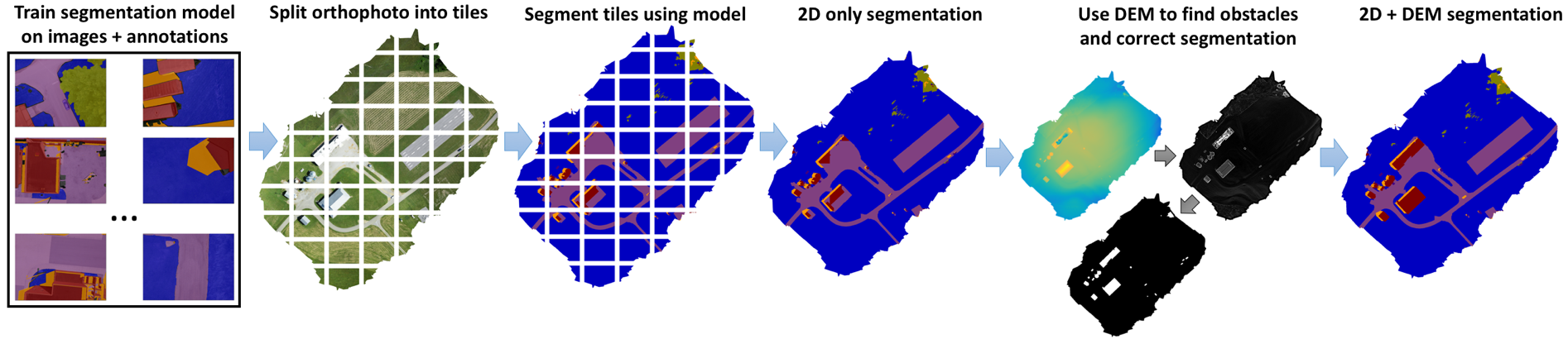}
    \caption{Overview of our approach to performing semantic segmentation of the aerial imagery. We first train a segmentation model on a dataset of images taken from low-flying UAV and their annotations.  We then take the orthophoto and divide it into tiles so that each can be segmented individually. These segmentations are then combined to make the 2D only segmentation result. To improve the segmentation results the DEM is then used to make updates. Regions surrounded by larger gradients are identified after which any pixels within those regions classified as traversable categories are assigned the mode of the most likely non-traversable categories within those regions.}
    \label{fig:segmentation_overview}
\end{figure*}

To perform semantic segmentation we use the Automatic Labeling Environment (ALE)~\cite{ladicky2011thesis}, which trains a model using 2D images and annotations and then uses that model to perform inference at the pixel-level on images in a test set unseen to the model.  The traversable categories are road and grass, and the rest are the non-traversable categories.  While shadows often contain traversable regions, we treat them as obstacles. It is possible to postpone analysis of those areas until a UGV enters the scene with LiDAR to analyze whether or not they are traversable, but we do not do this.  We identify obstacles in the DEM by calculating the gradient magnitude and filling regions surrounded by larger gradients.  Pixels within these regions that contain traversable category labels are assigned the mode of the most likely non-traversable categories within the region using the unaries computed by TextonBoost~\cite{shotton2006textonboost} in the ALE framework. An overview of our approach to performing semantic segmentation is shown in \figref{fig:segmentation_overview}.

\revieweraddition{
We note that we make corrections to regions of the segmentation where an obstacle has been detected and a traversable category has been classified, but do not make corrections to regions where no obstacle has been detected. We do this for two reasons: 1) grass and road are segmented with high precision, as evidenced by our results, and 2) there may be some obstacles that are not detected with the DEM. 
}

\section{Path Planning}
\label{sec:path_planning}

To plan a path for the UGV to visit points of interest on the ground we consider first how to plan a path between two points given an orthophoto, DEM, and segmentation.  Our method of choice was A$^*$~\cite{hart1968formal}, which extends Dijkstra's algorithm~\cite{dijkstra1959note} via a heuristic to assist in finding the best path between the two points.  The orthophoto, DEM, and segmentation have the same image dimensions, and we therefore define nodes of our graph to be the pixel positions with 8-pixel connectivity. The size of the grid paths are calculated on is 458x440 (201,520 total nodes), which translates to each pixel representing an area of approximately 0.6m x 0.6m. Downsampled versions of the orthophoto, DEM, and segmentation\footnote{The original dimensions for each of these outputs was 18137x17454.} were used to make calculating the paths efficient but still accurate. The cost function for A$^*$ search is defined as

\begin{equation}
f(n) = g(n) + h(n)
\end{equation}

where $g(n)$ is the cost of making a move between the current node ($x_c$) and a neighboring node ($x_n$), and $h(n)$ is the heuristic that estimates the cost of moving from $x_c$ to the goal node ($x_g$).  

\begin{figure} [h!]
    \centering
    \includegraphics[width=0.5\columnwidth]{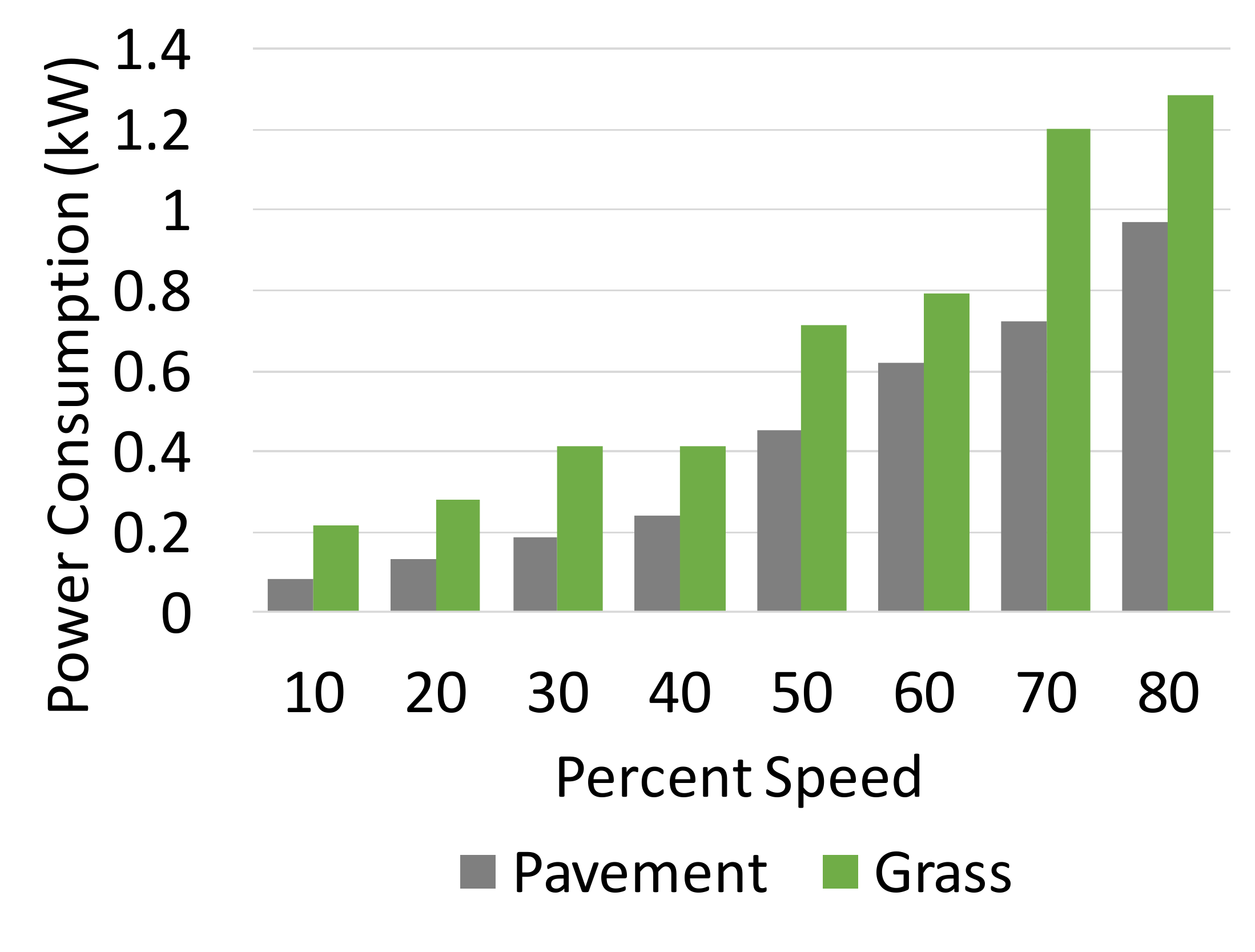}
    \caption{This shows the power consumption for the motors of the TURTLE when operating on pavement vs grass for different speed settings. The power consumption measurements were calculated by taking the median value of all the peaks in the plot over time.  Significantly less power is consumed on pavement compared to grass.}
    \label{fig:grass_vs_roads_turtle}
\end{figure}

Our implementation will find a path between two points in the orthophoto using the semantic segmentation results, where there is a preference that the path chosen follows the roads.  We experimented with using the DEM in the cost function, but found it made little difference, possibly because obstacles are not added as traversable nodes in the graph. However, for other scenes this cost can easily be included if necessary. The motivation for following roads over grass is that grass tends to be more difficult to traverse for UGV, as well as obstacles and ditches being less visible.  To provide further motivation for this design choice, we show the power consumption of the motors when traversing pavement and grass as a function of the percent speed set in \figref{fig:grass_vs_roads_turtle}. The power consumption values were calculated by taking the median value over all peaks in the plot over time as the UGV traversed both grass and pavement surfaces. As seen, traversing grass always results in a higher power consumption than when traversing pavement.

We calculate the heuristic function $h(x_c)$ as the euclidean distance in pixels to the goal position, and the cost of moving between $x_c$ and $x_n$ as


\begin{align}
g(n) = \wb^\intercal [\phi_1(x_c); \phi_2(x_n); \phi_3(x_c)].
\end{align}

\postrebuttal{The specific weights we use are $\wb=[5; 2; 5]$. Here we note that prior to the experiments we used different weights, with regular (not inverse) distances for $\phi_1(x_c)$ and $\phi_3(x_c)$, and a value of 1 in $\phi_2(x_c)$ when $x_n$ \emph{is} classified as road and 0 otherwise. We inverted the formulation so that negative weights were not used and observed \emph{very} similar planned paths to the ones presented in the paper. However, we have presented the weights and features with the correct formulation. The reason the planned paths may be similar are that they do not navigate around too many obstacles to reach the destination, and are still being rewarded for actions in a very similar way.} The following describes each feature $\phi_i$:

\begin{itemize}
\item $\phi_1(x_c)$ is the inverse of the distance to the nearest $x_i$ not classified as road, which is 0 when the $x_c$ is not classified as road. This rewards the algorithm for staying near the center of the road.
\item $\phi_2(x_n)$ is an indicator variable that is 1 when $x_n$ is not classified as road and 0 otherwise. This encourages the algorithm to pick roads over nodes classified as other categories.
\item $\phi_3(x_c)$ is the inverse distance to the nearest $x_i$ not classified as road or grass, which is 0 when $x_c$ is not classified as grass. This helps the path stay clear of obstacles in the scene.
\end{itemize}


\revieweraddition{The weights ($w$) used above are up to the system designer to pick for their application. With our current weights, there is a preference to follow roads over grass, but the algorithm does not minimize the amount of time spent on grass.
While we do not use weights based on the values of \figref{fig:grass_vs_roads_turtle}, this could be done with more data. This also extends to an arbitrary category size if traversability data can be gathered for each category. 
\postrebuttal{We did not use these costs in our experiments, because the choice between grass and roads were based on common sense reasoning that roads are better than grass. However, with other terrain types this might not be so clear, and so these types of plots may be very useful.}
Power consumption or other characteristics related to different terrain types being classified could be directly incorporated into these weights. These weights could also be actively modified as a UGV navigates a particular scene learning more about the environment.}

\section{Experiments}
\label{sec:experiments}

\subsection{Experimental Setup}

\paragraph{Radiation Sources.}
The sources used in the experiments are listed in \tableref{tab:rad_source_info}. Sources in these activity ranges are typically used for system checks of laboratory equipment.  As such they are relatively weak and sit roughly at the threshold of detection for the detector systems used. The sources were placed in Nalgene bottles and positioned on top of thin steel stands 1m off the ground for the aerial data collection and taped to the bottom of the stands for the ground collection to ensure that there was no attenuation from the stand during the ground-based measurements.

\begin{table}[h!]
\caption{Information for each of the 4 radiation sources used in the experiments. Different combinations of these sources are used when creating each source location.} 
\label{tab:rad_source_info}
\begin{center}
\begin{tabular}{ccc}
Nuclide & Half-life (yr) & Activity ($\mu$Ci)\\ \hline \hline
$^{137}$Cs & 30.2 & 10.0 \\
$^{133}$Ba & 10.7 & 16.1 \\
$^{166m}$Ho & 1200.0 & 138.7 \\
$^{166m}$Ho & 1200.0 & 147.1 \\
\hline
\\ 
\end{tabular}
\end{center}
\end{table}

\paragraph{RMAX Missions.}
Two flights, using the RMAX, were conducted at Kentland Farms, Blacksburg, VA, where different configurations of radiation sources were placed in the scene for each flight. In Mission 1, all of the sources listed in \tableref{tab:rad_source_info} were placed at a single location.  In Mission 2, both Ho sources were placed at one location, and the Ba and Cs sources were placed at another location. 
\revieweraddition{
Source locations in each mission are estimated by the GPS locations associated with the maximum counts (sum of the 1024-d radiation signal from the detector).
}

\begin{figure}[h!]
    \centering
    \includegraphics[width=0.4\columnwidth]{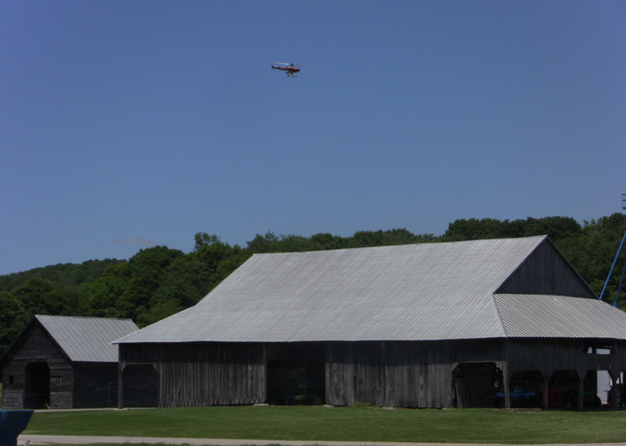}
    \caption{The Yamaha RMAX mid-flight during the first search mission.}
    \label{fig:yamaha_rmax}
\end{figure}

For each flight, paths were generated in Mission Planner~\cite{missionplanner}, with an altitude of 30m, and a distance between scan lines of 4m, which was chosen to ensure sufficient overlap in the imagery for generating a high quality 3D reconstruction and also to obtain more dense measurements for the radiation data. \gordonfirst{The height of 30m allowed testing of the detection system's capabilities given the low activity level of the sources with a significant background signature present.} The velocity of the RMAX was set to 3m/s, and images were captured once a second, resulting in around a 90\% overlap for subsequent images. A total of 1644 images (874 for Mission 1 and 770 for Mission 2) were collected for both missions, with 3288 images if considering stereo pairs. Image taking is also synchronized to simultaneously log GPS and radiation data. The RMAX is seen flying one of the missions in \figref{fig:yamaha_rmax}. While the altitude was chosen based on observations of the scene to ensure there would be no hazard for the RMAX during the flight, this value could be chosen using a 3D point cloud of the scene. If one is not available, then real-time sense-and-avoid methods could be applied.

\revieweraddition{
Mission 1 lasted approximately 23 minutes, and Mission 2 lasted approximately 26 minutes. This includes take off, landing, the navigation to and from the start/end waypoints. Agisoft required several days of computation, while segmentations were finished in a matter of seconds per image. We performed the UGV missions at a later date than the flights due to logistics. Moving forward we are exploring methods to obtain 3D reconstructions much faster by using stereo vision so that this is not necessary.
}

\revieweraddition{
The reason that we did not survey the entire area at once is because, 1) we wanted to conduct multiple missions, and 2) the endurance of the RMAX may not have been sufficient. The RMAX can fly for approximately 45 minutes before needing to refuel. If endurance is needed, then a fixed wing aircraft is better, but may move too quickly to collect statistically meaningful data. There are higher endurance helicopters, but they are much more expensive than the RMAX, which was within our budget. Our requirement to be able to carry over 30 lbs of payload also narrowed the helicopter selection.
}

\paragraph{TURTLE Missions.}
We ran two separate missions for each configuration of radiation sources.  The destinations are defined by the position of max counts from the aerial radiation data. The same start position is set for both missions, which is located on one of the roads at the exterior of the scene. Since the scene may change between when the flights take place and when the TURTLE is deployed, we placed an obstacle on the planned path for the TURTLE in both missions so that it was forced to detect the obstacle and then find a path around it by updating the global map and planning an alternative route. \gordonpratap{The global map is updated by removing nodes in the 2D grid containing the obstacle from the set of traversable nodes.}

\subsection{\gordonfirst{RMAX Results.}}
\label{sec:experiments_rmax}

\begin{figure*}[h!]
    \centering
    \subfigure[Aerial search path for the first configuration, where 4 radiation sources (2 Ho, 1 Ba, 1 Cs) are placed at a single location.]{
		\label{fig:flight_paths_1}
		\includegraphics[width=0.48\columnwidth]{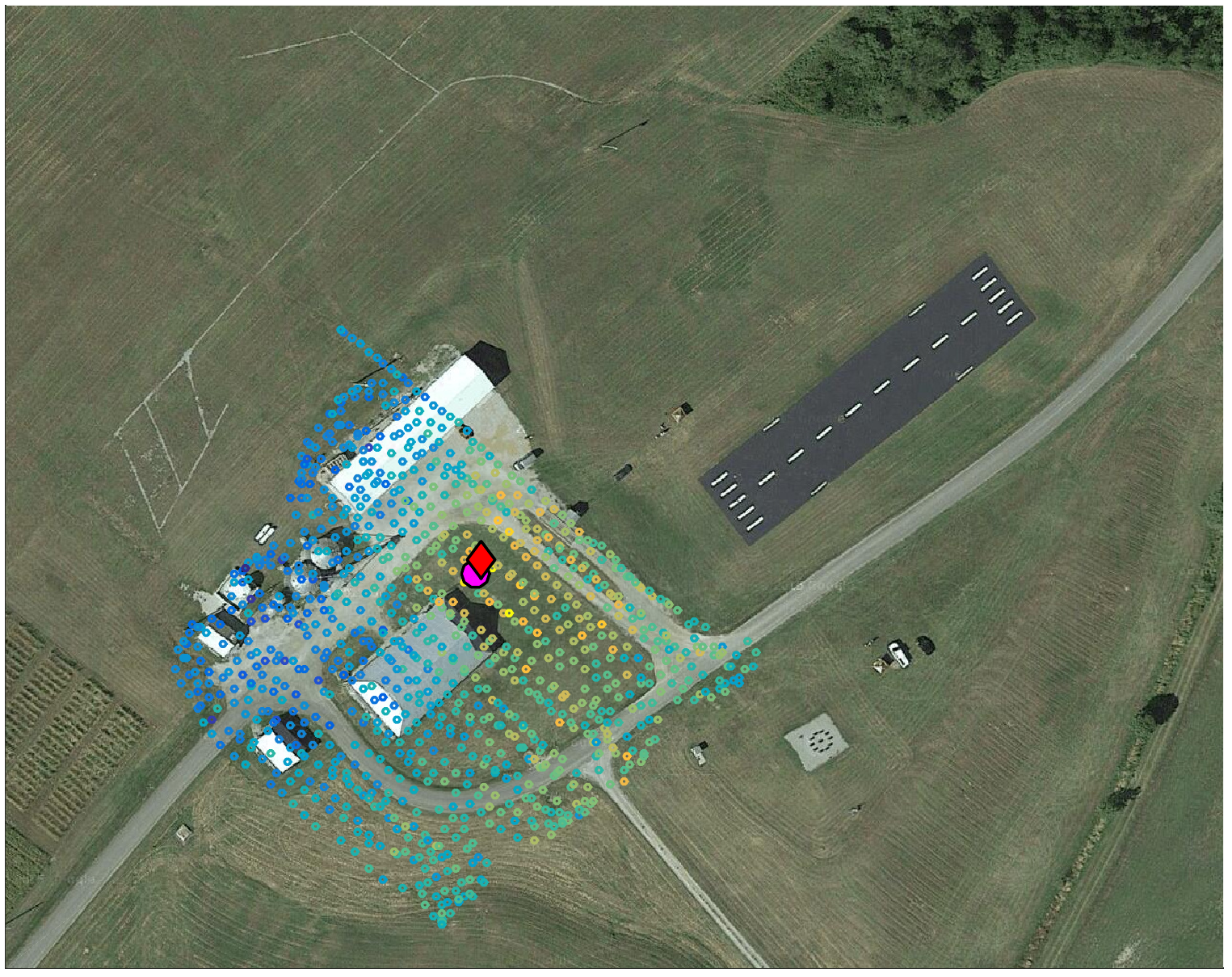}
	}
	\subfigure[Aerial search path for the second configuration, where 2 Ho sources are placed at one location (position closest to the location of max counts), and 1 Ba and 1 Cs sources are placed at a second location.]{
		\label{fig:flight_paths_2}
		\includegraphics[width=0.48\columnwidth]{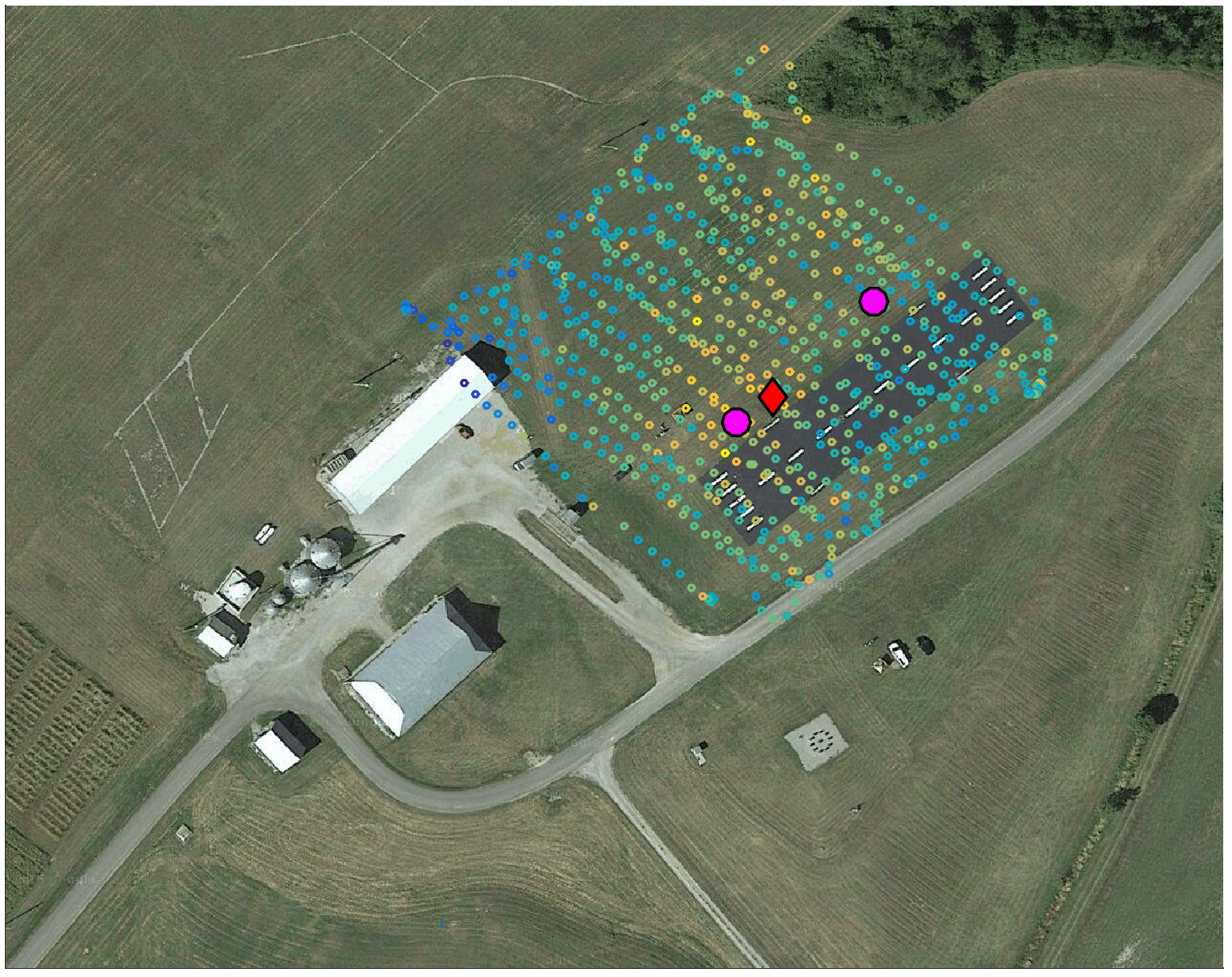}
	}
    \caption{The first (a) and second (b) flight paths at Kentland Farms, Blacksburg, VA, shown in Google Maps, where the color of each point represents the counts, calculated by summing the 1024-d spectral vector at each position.  The magenta circles show the ground truth locations of the sources, and the red diamonds show the positions of max counts, which are set as destinations for the UGV to visit and take additional measurements.}
    \label{fig:flight_paths}
\end{figure*}

\paragraph{Radiation Results.}
The position histories for each flight are shown in \figref{fig:flight_paths}, where the color of each point represents the counts value (from blue/low to red/high). The ground truth locations of the sources are shown with magenta circles, and the red diamonds show the positions of max counts set as destinations for the TURTLE. In our results we are successfully able to identify 2 of the 3 source locations of the combined missions. The failure case is the Ba and Cs source combination, as seen in \figref{fig:flight_paths_2} as the ground truth position farthest from the max counts estimate. It was found to be too weak to be seen by different nuclear anomaly detection algorithms at the altitude flown by the RMAX (30m). The counts value at the closest reading was 612, and the median of the 10 closest measurements was 617.5, both of which were below the average counts for all of the aerial readings taken during that flight. For reference, the counts for the position closest to the other source location (2 Ho sources) is 654, with a median for the closest 10 points of 658. Therefore the TURTLE is never instructed to visit anywhere near this position unless the starting point is set in such a way that it passes right by it on the way to the location of the 2 Ho sources, which is much stronger. 

\revieweraddition{
While our experiments proved what we set out to prove, the failure case does provide motivation for UGV-based search methods to be applied. 
The particle filter method presented for aerial search in~\cite{kochersberger2014post} is one example of an approach that could be applied for ground search operations. Other approaches, such as maximum-likelihood estimation (MLE) and contour following~\cite{towler2012radiation}, also have potential. This work is part of a fundamental research project, so there are currently no specific end-user requirements for the UAV-UGV teaming side of our work. Future work may include coming up with performance metrics to evaluate the performance of more advanced radiation search tasks.
}

\begin{figure*}[h!]
    \centering
    \subfigure[Mission 1. Background flight: $\mu$ = 558, $\sigma$ = 38.9. Source flight: $\mu$ = 606.7, $\sigma$ = 48.1.]{
        \label{fig:mission1_counts_hist}
        \includegraphics[width=0.48\columnwidth]{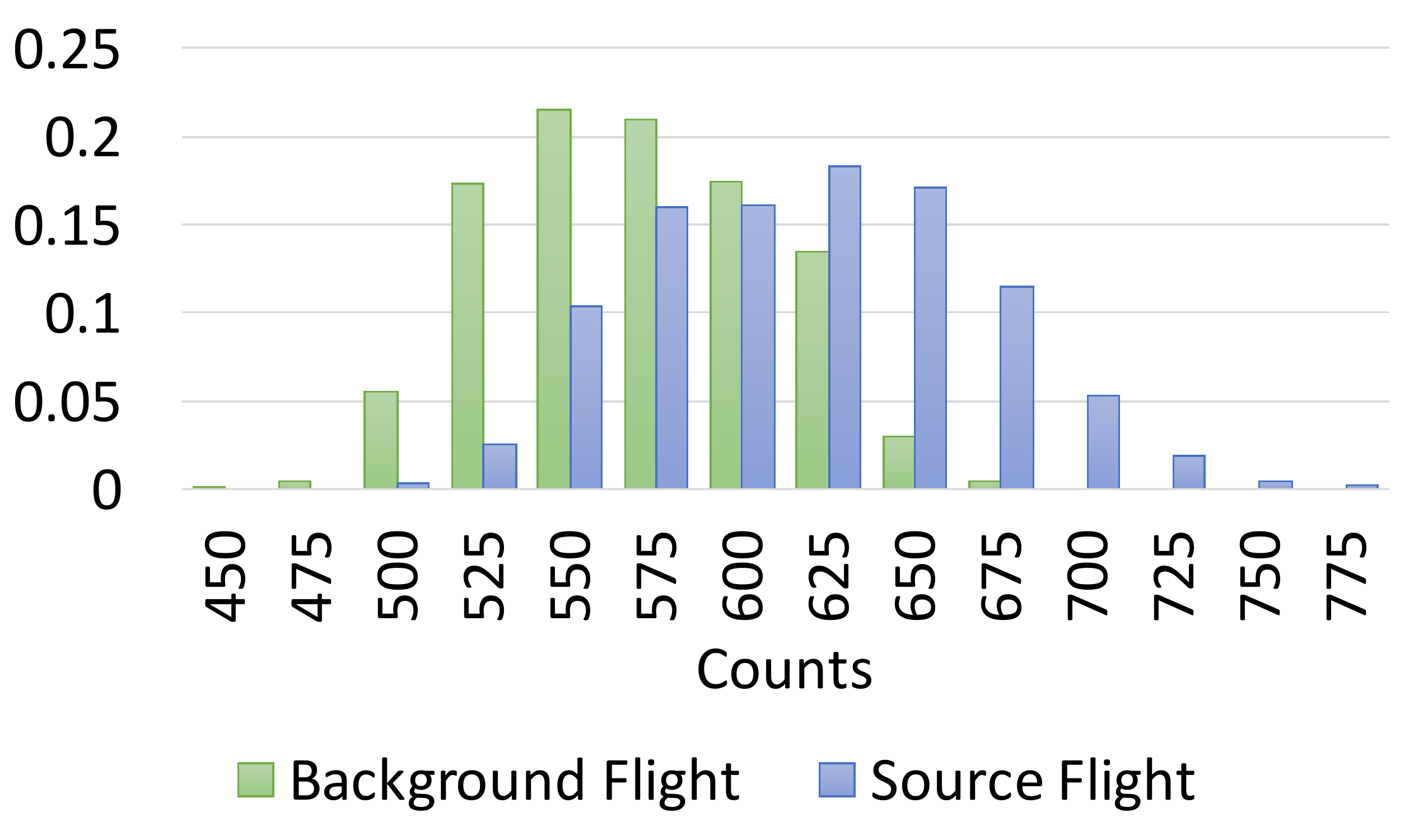}
    }
    \subfigure[Mission 2. Background flight: $\mu$ = 593.9, $\sigma$ = 30.9. Source flight: $\mu$ = 617.6, $\sigma$ = 33.4.]{
        \label{fig:mission2_counts_hist}
        \includegraphics[width=0.48\columnwidth]{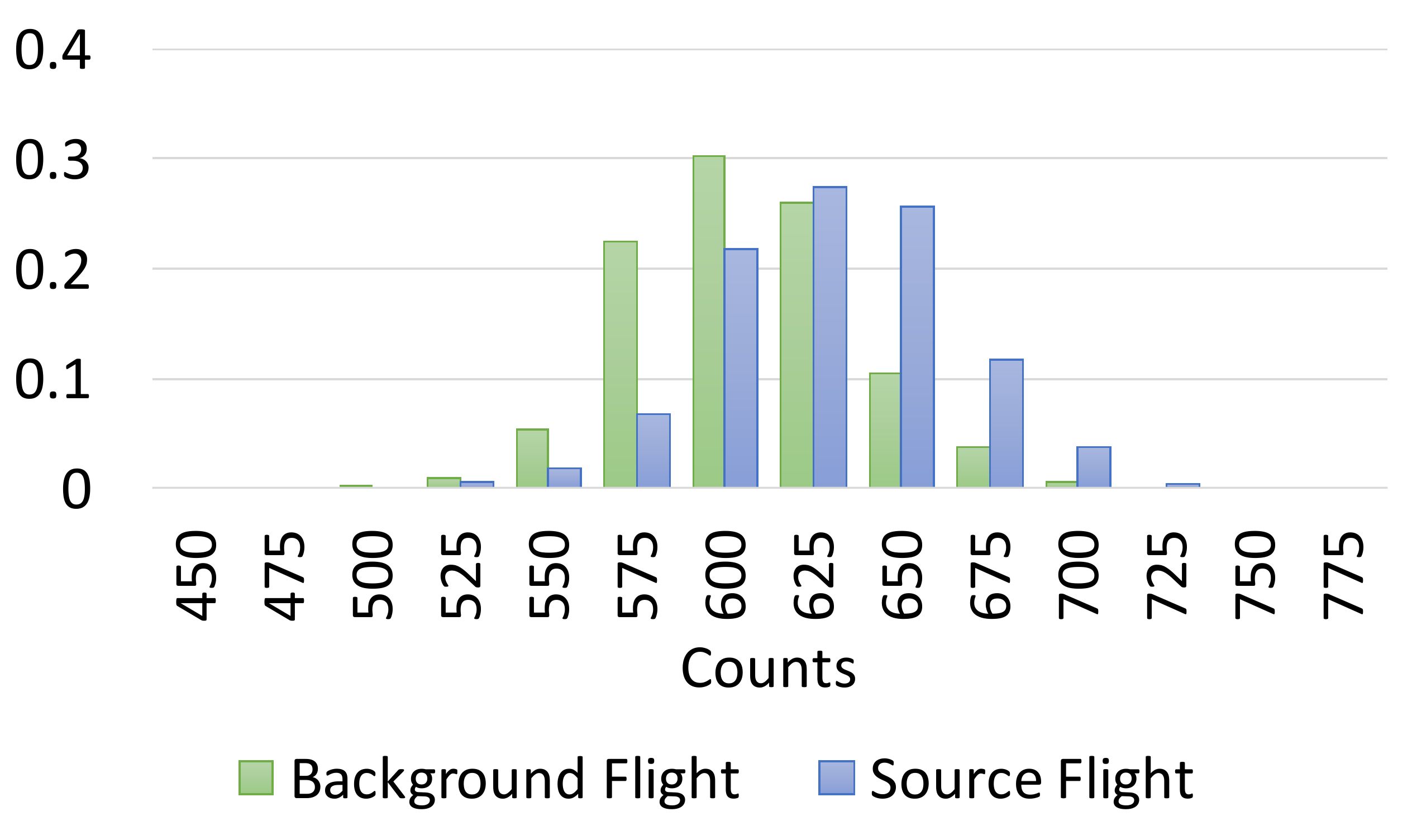}
    }
    \caption{Histograms of the counts for each mission (normalized), which includes the main mission with radiation sources and the background scans. For each mission we ran $t$-tests between the counts for the background and source flights to verify that statistically significant differences were observed. In both cases reject the null hypothesis, that their means are identical, with a p-value of 0.05.}
    \label{fig:missions_counts_hists}
\end{figure*}

For each mission, we performed a background scan of the mission area and a flight for the main mission with radiation sources present. These background scans are never used to assist in finding the sources, but help provide context for the data observed during the source flights. For each mission we ran paired $t$-tests between the background and source flights for each mission to test the null hypothesis that the counts (not normalized) have identical means. In both cases we were able to reject this null hypothesis with a p-value of 0.05, and conclude that statistically significant observations were made during the source flights. Histograms of the background and source flights for each mission are shown in \figref{fig:missions_counts_hists}.

\begin{figure*}[ht!]
    \centering
    \includegraphics[width=\textwidth]{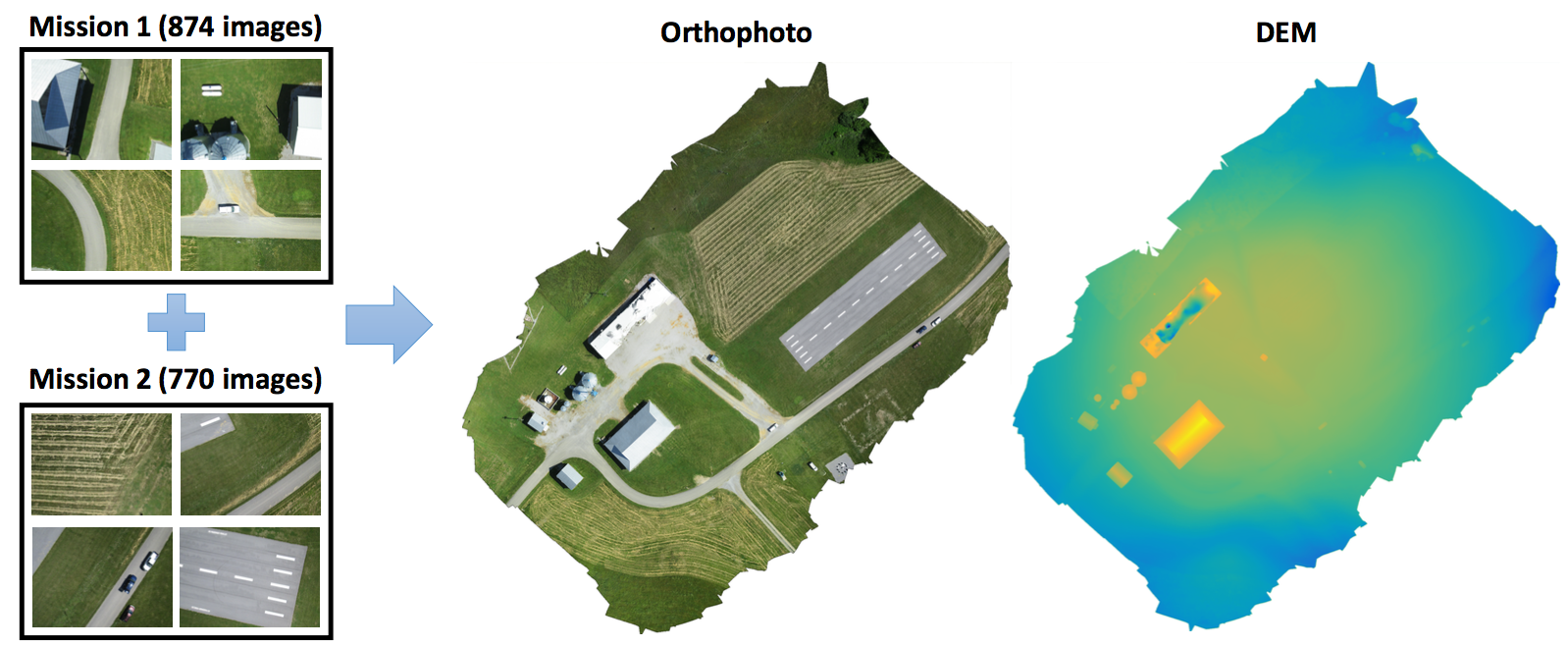}
    \caption{The orthophoto and DEM generated by Agisoft.}
    \label{fig:ortho_dem_google}
\end{figure*}

\begin{figure}[h!]
    \centering
    \includegraphics[width=0.35\columnwidth]{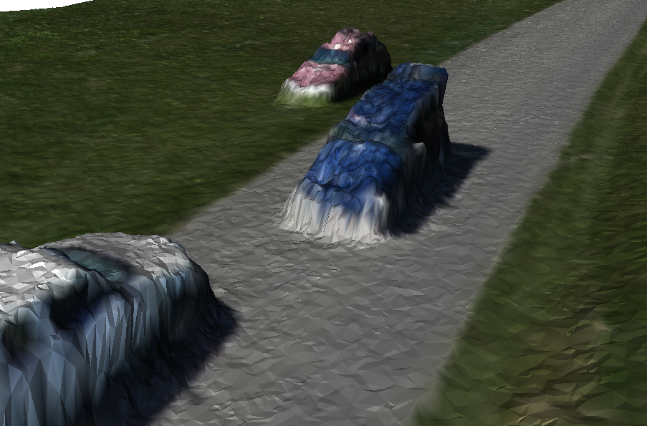}
    \caption{A view of the point cloud of Kentland Farm generated using Agisoft that illustrates the level of detail possible using off-the-shelf cameras.}
    \label{fig:up_close_3d}
\end{figure}

\paragraph{Orthophoto and DEM.}
The orthophoto and DEM output by Agisoft are shown in \figref{fig:ortho_dem_google}. Note that the DEM values are incorrect for the building with the white roof. This does not affect path planning, however, as this area can still be identified as non-traversable because of the discontinuity with surrounding regions.  Also, this provides further motivation for the 2D semantic segmentation of the aerial images. A close up of vehicles is shown in \figref{fig:up_close_3d} to illustrate the level of detail in the final output by Agisoft.

\revieweraddition{While GPS was used for all experiments presented in this paper, the approach has the potential to work in GPS-denied areas, where the mission is still feasible, but certainly more difficult. SfM and stereo vision can still be used to generate a global map, but the individual systems must now operate by transforming their local coordinates to shared global coordinates. One example of how this might work would be to have the UGV identify landmarks that can be matched to locations in the aerial map generated by the UAV, and then use these positions to align the two maps. Generating aerial maps (SfM, stereo, \emph{etc.}) without GPS can mean longer run times and less accurate reconstructions. For example, part of the SfM pipeline includes features being matched between pairs of images when generating the 3D reconstruction. A naive approach would search for matches in ${\#images \choose 2}$ pairs of images. GPS can be used to only search for matches in pairs of images located near one another. Without GPS, we can still limit the number pairs to be searched by image clustering (\eg clustering using GIST image descriptors \cite{oliva2001modeling}). While not as robust as using GPS, and still resulting in pairs of images with no matches, this is still much faster than the naive approach.}

\revieweraddition{
When dealing with uncertainty in the GPS measurements, algorithms such as Kalman filters can be used with visual SLAM to update position beliefs. One concern for our application is having an inaccurate path be sent to the UGV. While this is not ideal, the UGV would still be able to scan the terrain around it to determine what is traversable and what is not. To correct the path, identifying landmarks that are matched to the aerial map to make the correction would be one possibility. Another possibility would be to have the UGV perform semantic segmentation to make the correction. For example, the GPS coordinates output for a path along a road may be located on a neighboring grass region. Semantic segmentation would allow the UGV to identify the location of the road nearby and make the correction.}

Another problem with using a single-camera system in GPS-denied environments is the inherent scale ambiguity associated with structure from motion.
Our 2-camera imaging system can be used to resolve this~\cite{christie2014spie}, by scaling the 3D reconstruction to an interpretable size using the known baseline between the cameras. A UGV could then localize itself in a local coordinate frame defined by the 3D reconstruction.

\paragraph{Semantic Segmentation and Path Planning.}
We perform semantic segmentation on tiles of the orthophoto using ALE~\cite{ladicky2011thesis} and then compare our results to ground truth annotations.
Per-category results are shown in \tableref{tab:ss_results}.  We measure results in terms of precision and recall, where for each category $c$ precision calculates how many of the instances classified as $c$ are correct, while recall calculates how many of the ground truth instances labeled $c$ have been correctly classified. True positives (TP), false positives (FP), and false negatives (FN) are used to calculate precision and recall as

\begin{equation}
\text{precision} = \frac{\text{TP}}{\text{TP}+\text{FP}}, \text{    recall} = \frac{\text{TP}}{\text{TP}+\text{FN}}.
\end{equation}

\begin{figure*}[h!]
	\centering
	\subfigure[2D + DEM confusion matrix.]{
		\label{fig:confusion_matrix1a}
		\includegraphics[width=0.45\columnwidth]{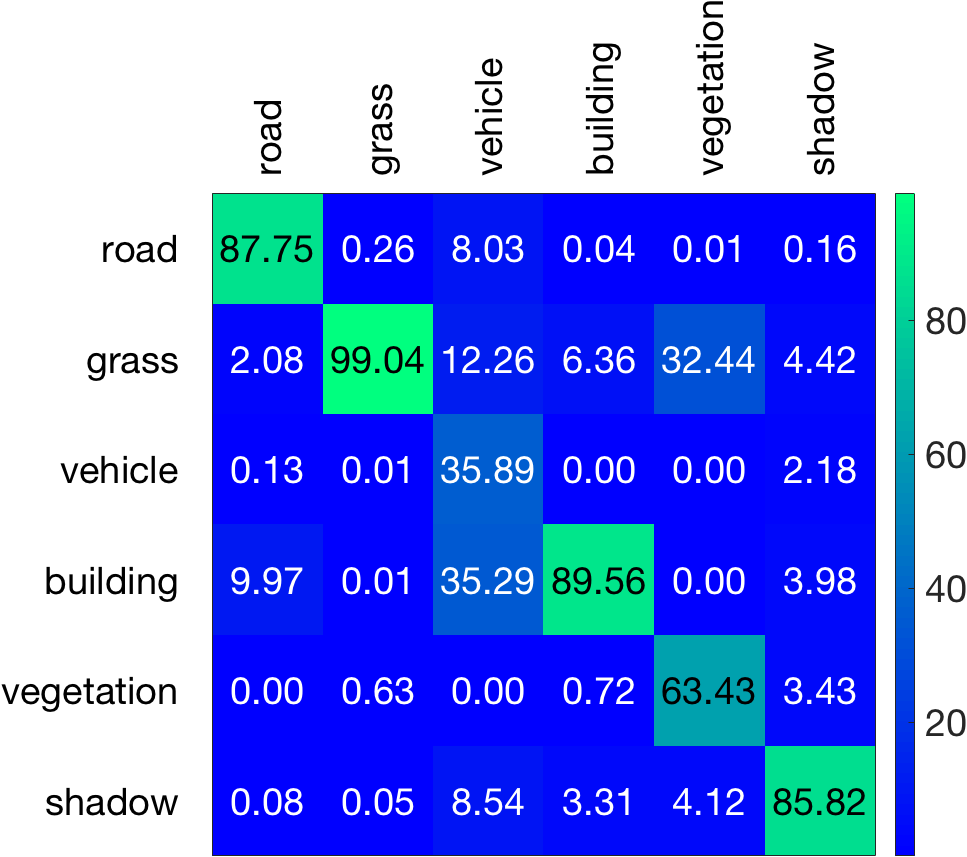}
	}
	\quad
	\subfigure[2D only confusion matrix]{
		\label{fig:confusion_matrix1b}
		\includegraphics[width=0.45\columnwidth]{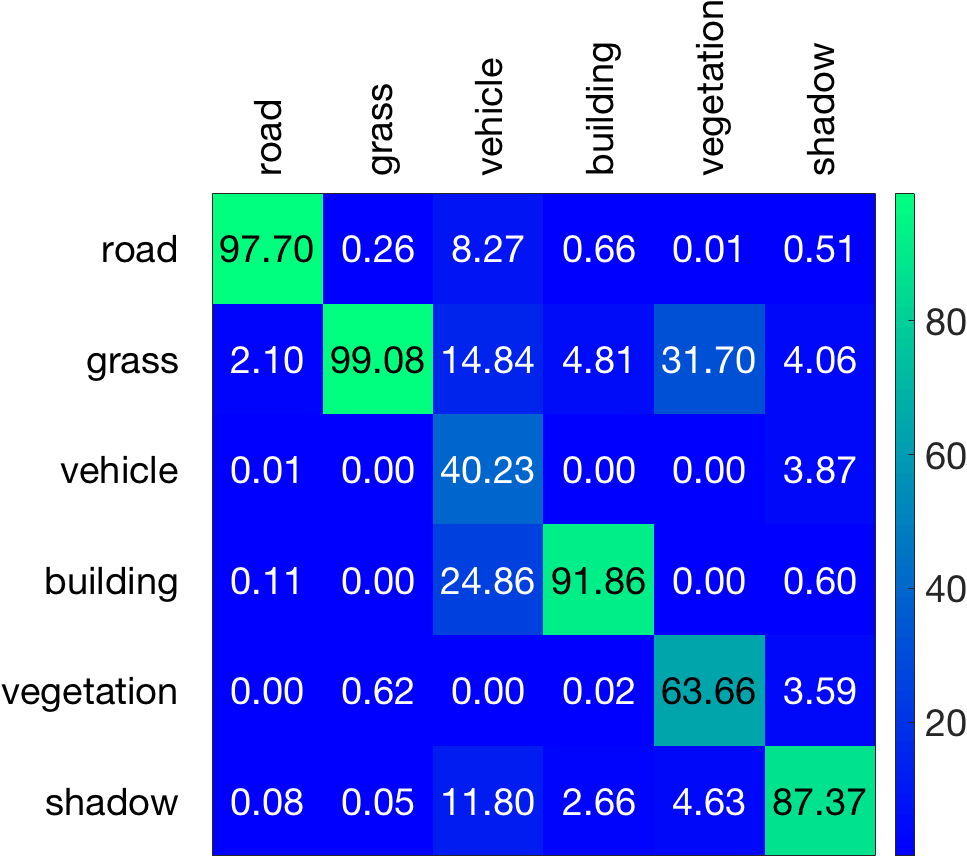}
	}
	\caption{The confusion matrices for both our approach of using the orthophoto and DEM to perform semantic segmentation, and a 2D only baseline that only uses the orthophoto. The diagonal elements of the confusion matrices show the precision values \revieweraddition{from \tableref{tab:ss_results}. The different colors in confusion matrices represent values between 0 and 100.}}
    \label{fig:confusion_matrix}
\end{figure*}

\begin{figure*}[h!]
    \centering
    \subfigure[Ground truth image.]{
		\label{fig:gt_kentland}
		\includegraphics[width=0.37\columnwidth]{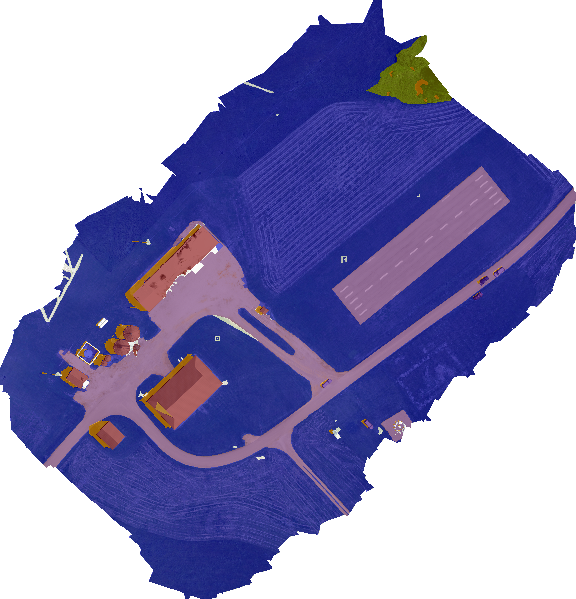}
	}
	\hspace{.15in}
	\subfigure[Segmentation result.]{
		\label{fig:ale_seg}
		\includegraphics[width=0.37\columnwidth]{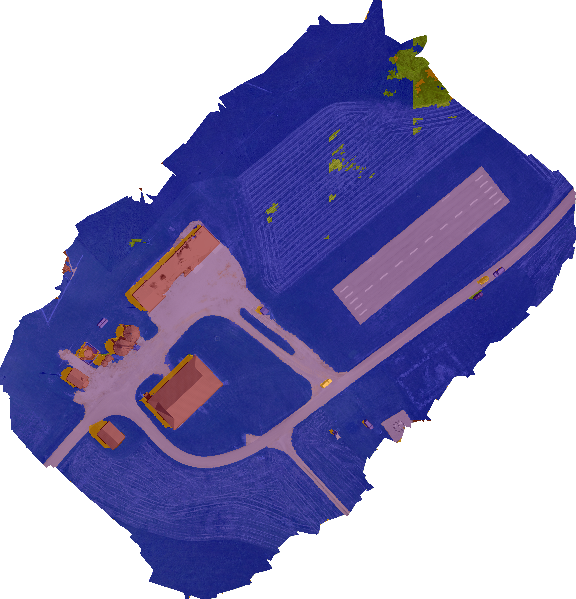}
	}
    \hspace{.15in}
    \includegraphics[height=4.55cm]{figs/legend}
    \caption{(a) Ground truth image of the orthophoto of Kentland Farms done with LabelMe~\cite{labelme2007}. (b) Result of segmenting the orthophoto by training the ALE~\cite{ladicky2011thesis} on our dataset and then refining the results using the DEM. See the legend in \figref{fig:aerial_dataset_overview} to map colors to categories.}
    \label{fig:segmentation_overlays}
\end{figure*}

We show that our approach of reasoning about the 2D orthophoto and DEM to output final predictions performs better than a baseline that only reasons about the 2D orthophoto.  The road and grass categories have very high accuracy, which is expected given that they are usually visually distinguishable from the other categories and each other. When confused with non-traversable categories, the DEM can be used to make corrections. The reliability of the model to segment these categories is also important for path planning, as these are the traversable categories for the UGV.
\gordonfirst{The confusion matrices for the results of our approach and the baseline are shown in \figref{fig:confusion_matrix}.}  Note that the non-traversable categories are typically confused with one another. This makes no difference for the path planner, but there is still motivation to improve performance on these categories as this is useful for high-level reasoning, such as understanding that a radiation source is more or less likely to be present at certain coordinates (\eg inside a vehicle). The ground truth annotation and semantic segmentation result are shown in \figref{fig:segmentation_overlays}.

\begin{table*}[h!]
\caption{Quantitative results for the semantic segmentation of the Kentland Farms imagery, showing per-category, average, and global accuracies for our approach (2D + DEM) that uses the orthophoto and DEM to reason about category prediction, and a 2D only baseline.} 
\label{tab:ss_results}
\begin{center}
\begin{tabular}{c|cccccc|cc}
method / metric & road & grass & vehicle & building & vegetation & shadow & Global & Average \\ 
\hline
\hline
2D precision & 87.75 & 99.04 & 35.89 & 89.56 & 63.43 & 85.82 & - & 76.92\\ 
2D + DEM precision& \textbf{97.70} & \textbf{99.08} & \textbf{40.23} & \textbf{91.86} & \textbf{63.66} & \textbf{87.37} & - & \textbf{79.98}\\ 
\hline
2D recall & \textbf{98.57} & \textbf{98.78} & 55.22 & 42.89 & 60.68 & \textbf{85.42} & 96.20 & 73.59\\ 
2D + DEM recall & 98.41 & 98.74 & \textbf{61.96} & \textbf{97.85} & \textbf{62.29} & 81.06 & \textbf{97.89} & \textbf{83.39}\\ 
\end{tabular}
\end{center}
\end{table*}

\begin{figure*}[ht!]
    \centering
    \subfigure[Planned path for Mission 1.]{
		\label{fig:planned_path_1}
		\includegraphics[width=0.4\columnwidth]{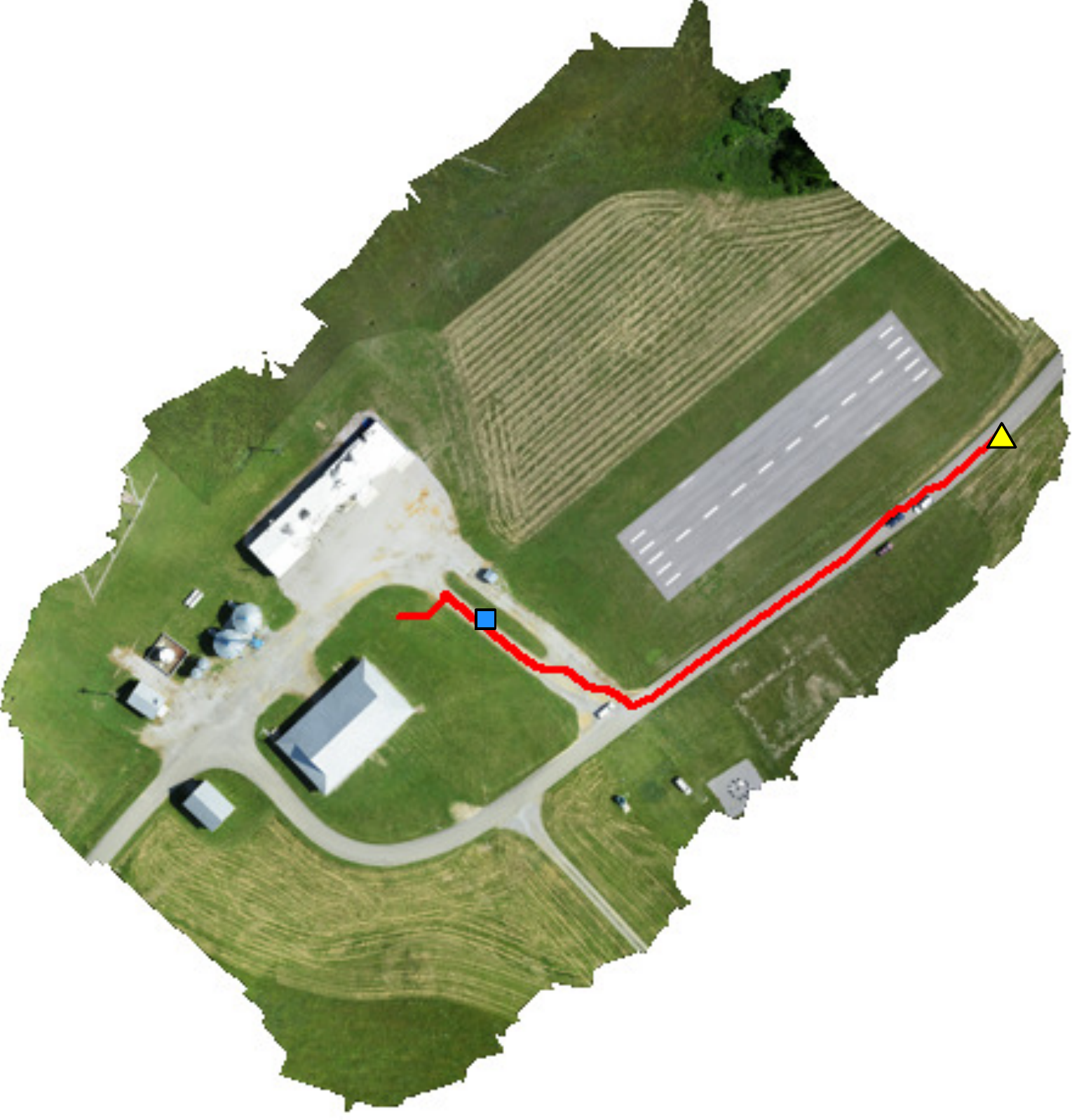}
	}
	\subfigure[Planned path for Mission 2.]{
		\label{fig:planned_path_2}
		\includegraphics[width=0.4\columnwidth]{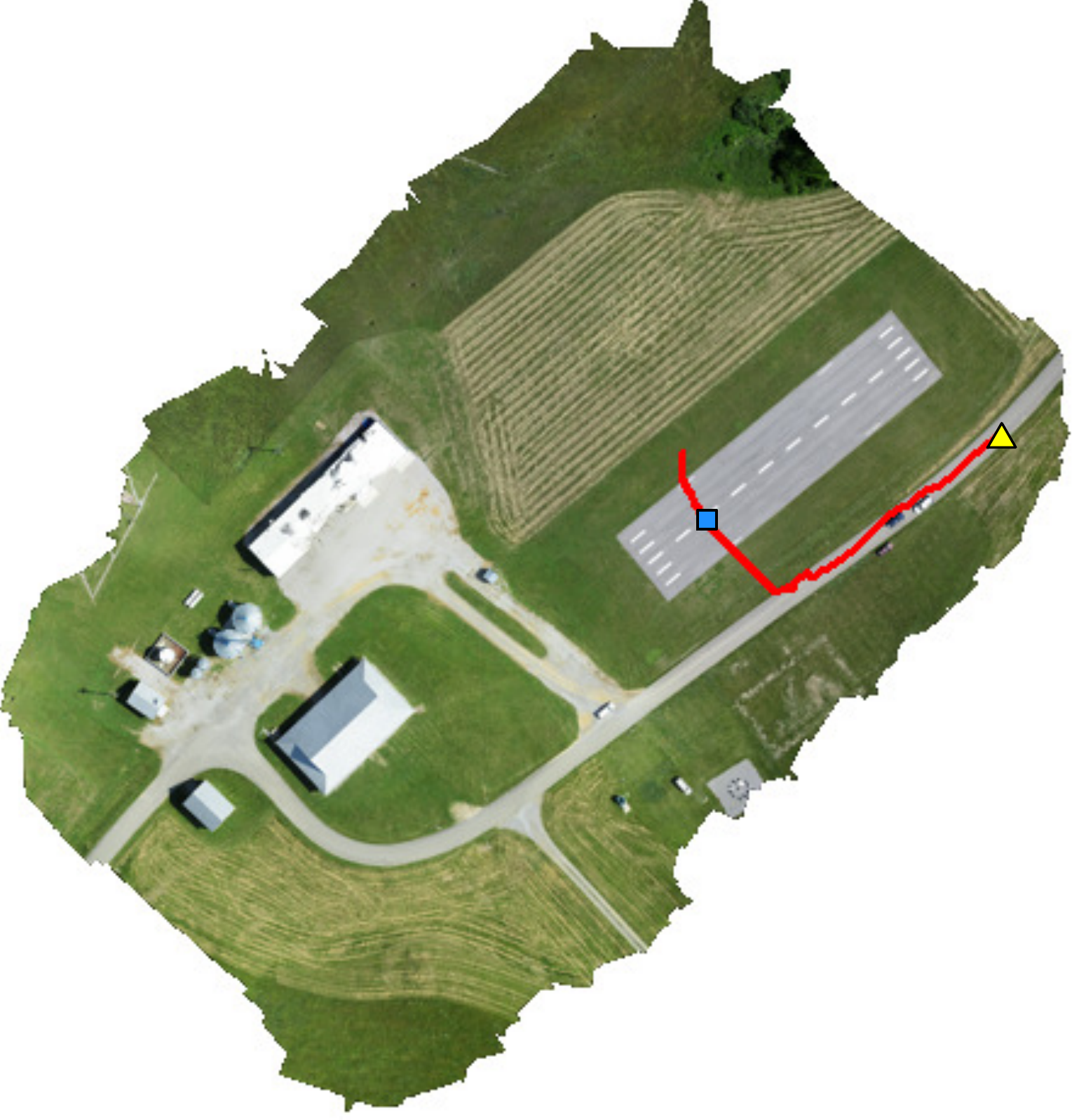}
	}
    \caption{The planned paths for each of the two radiation source configurations. The start position (yellow triangles) was set on the exterior points on the orthophoto containing a road. The blue square shows the position where an obstacle was placed so the TURTLE was forced to find an alternative path when encountered. These paths were each generated in a matter of seconds.}
    \label{fig:ugv_planned_paths}
\end{figure*}

The planned missions are shown in \figref{fig:ugv_planned_paths}, where the red pixels display the path, the blue squares shows the locations of the obstacles, and the yellow triangles show the start/end positions. As seen, the planned path plans around the vehicles on the road that were present during the flight, but not during the ground experiments.  
\revieweraddition{In our experiments, we do not update the global map to remove obstacles that were present in the aerial map, but are no longer in the scene. This was not necessary for our experiments, but note that this could easily be incorporated by adding an additional process to analyze the LiDAR data.}

\subsection{TURTLE Results}

\begin{figure*}[ht!]
    \centering
    \subfigure[LiDAR point cloud for source configuration 1]{
    	\label{fig:lidar1}
    	\includegraphics[width=0.48\columnwidth]{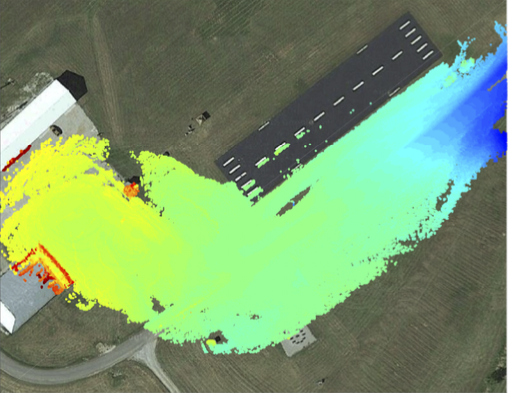}
	}
	\subfigure[LiDAR point cloud for source configuration 2]{
		\label{fig:lidar2}
		\includegraphics[width=0.48\columnwidth]{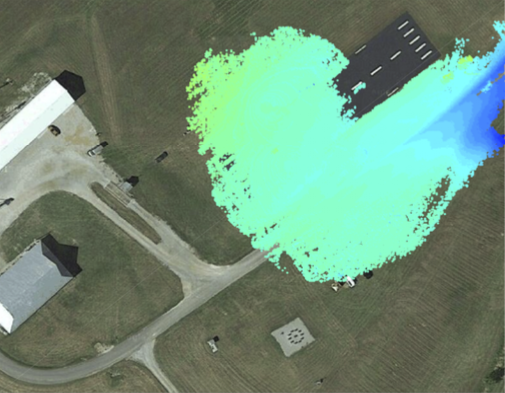}
	}
    \caption{Global DEMs generated by the TURTLE's LiDAR for each search mission. During the construction of the DEM, height values were averaged for points with the same ($x$,$y$).}
    \label{fig:lidar_maps}
\end{figure*}

The global LiDAR maps (DEMs) generated for each mission\footnote{These can be used in post processing to help understand the scene around the area of radiation activity.} by the TURTLE are shown in \figref{fig:lidar_maps}. When multiple height values are observed at the same ($x$,$y$) the values are averaged. This was done for efficiency reasons, as storing all previous values so that the $n$th percentile can be calculated requires a significant amount more storage. We also experimented with taking the max, but observed that this was susceptible to noise.  

\begin{figure*}[ht!]
    \centering
    \subfigure[Path taken for Mission 1.]{
		\label{fig:taken_path_1}
		\includegraphics[width=0.48\columnwidth]{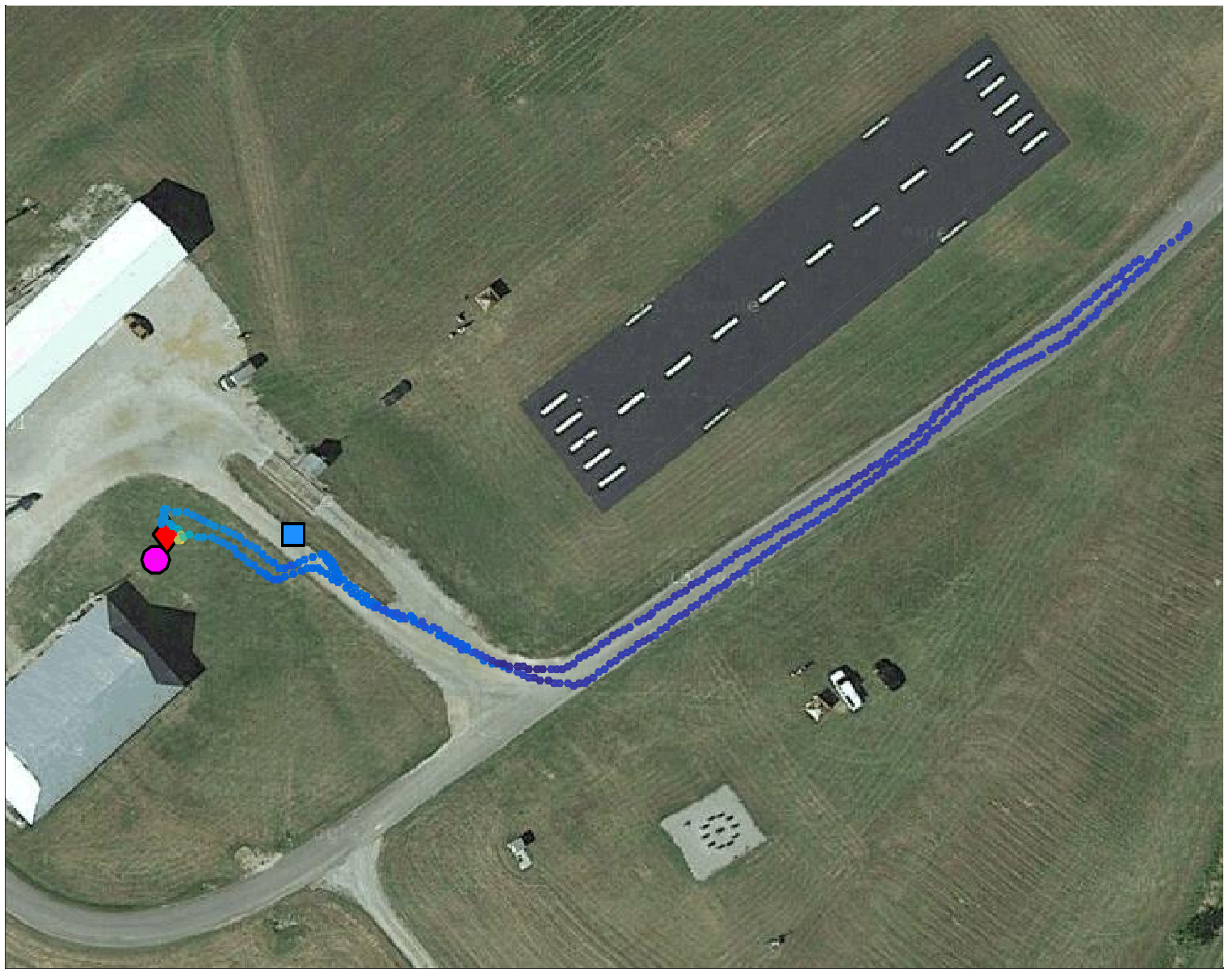}
	}
	\subfigure[Path taken for Mission 2.]{
		\label{fig:taken_path_2}
		\includegraphics[width=0.48\columnwidth]{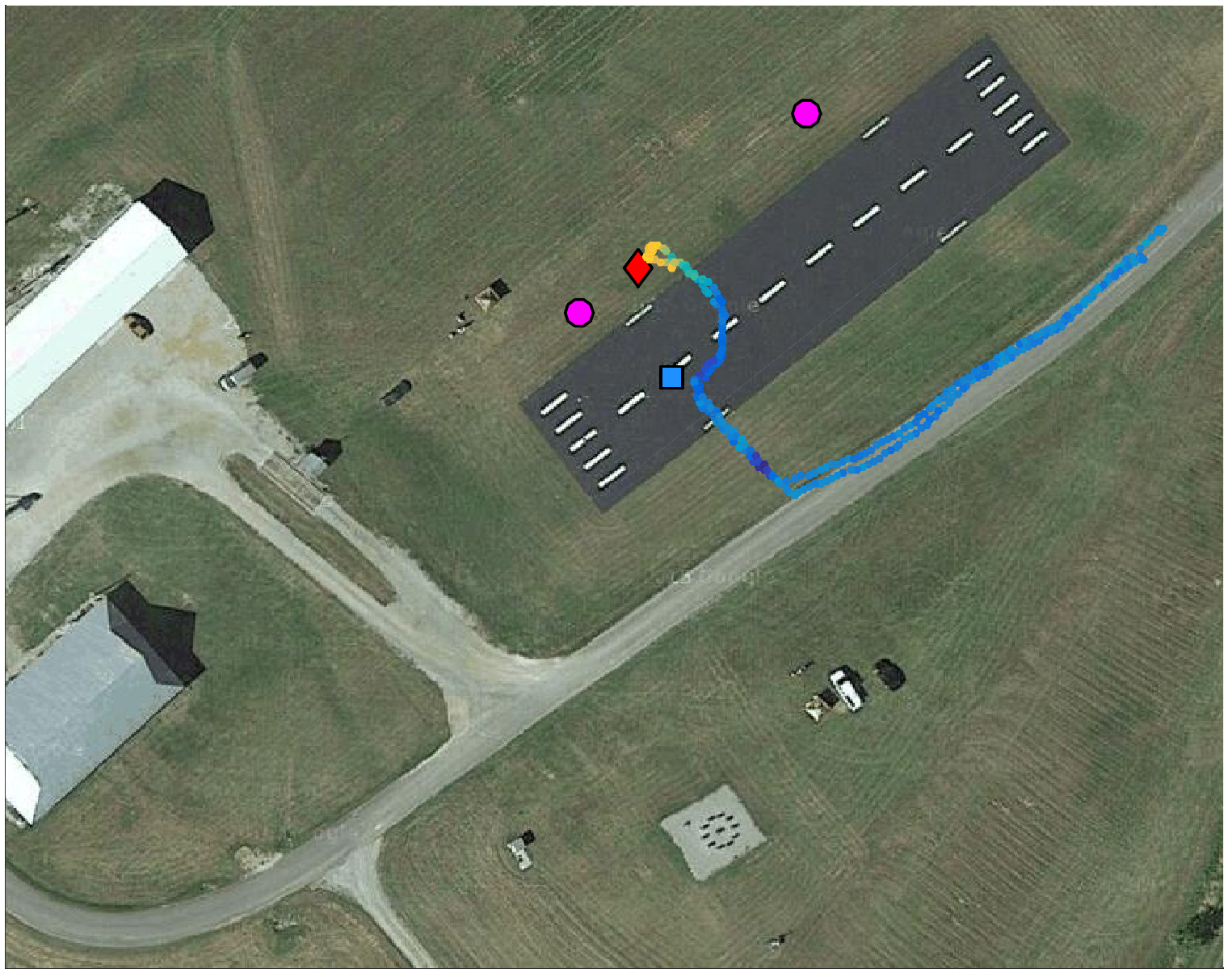}
	}
    \caption{Paths taken for the missions of both source configurations where the counts were used to map to the colors seen at each waypoint. The magenta circles show the ground truth locations of the two source positions, the red diamond shows the position of max counts from the aerial data, and the blue square shows the position of where the obstacle was placed. As seen in both missions, the TURTLE avoids the obstacles, which was done by reasoning with both local and global information.}
    \label{fig:ugv_final_paths}
\end{figure*}

\revieweraddition{
\textbf{Obstacle Detection and Avoidance.}
Local LiDAR scans were analyzed to find obstacles on or near the current path, and were used to update the global DEM and segmentation. Specific pixels associated with the obstacle are set in the segmentation, the region of which is dilated as a cautionary measure to make sure the full size of the obstacle is contained within the region that defines it in the segmentation. An updated path is then generated using the same path planning algorithm by taking the current position of the TURTLE as the start position, using the same goal position, and using the updated segmentation.
}
The final paths taken by the TURTLE, with obstacles avoided, are shown in \figref{fig:ugv_final_paths}. As seen, at each position in the path history the counts were mapped to a color value to represent intensity.  The difference from \figref{fig:ugv_planned_paths} can be seen where it has identified the obstacle and navigated around it. To return to the start position, we simply keep track of the waypoints visited on the way to the destination and then follow them back. 

\begin{figure*}[h!]
    \centering
    \subfigure[Source configuration 1]{
    	\label{fig:counts_config1}
    	\includegraphics[width=0.48\columnwidth]{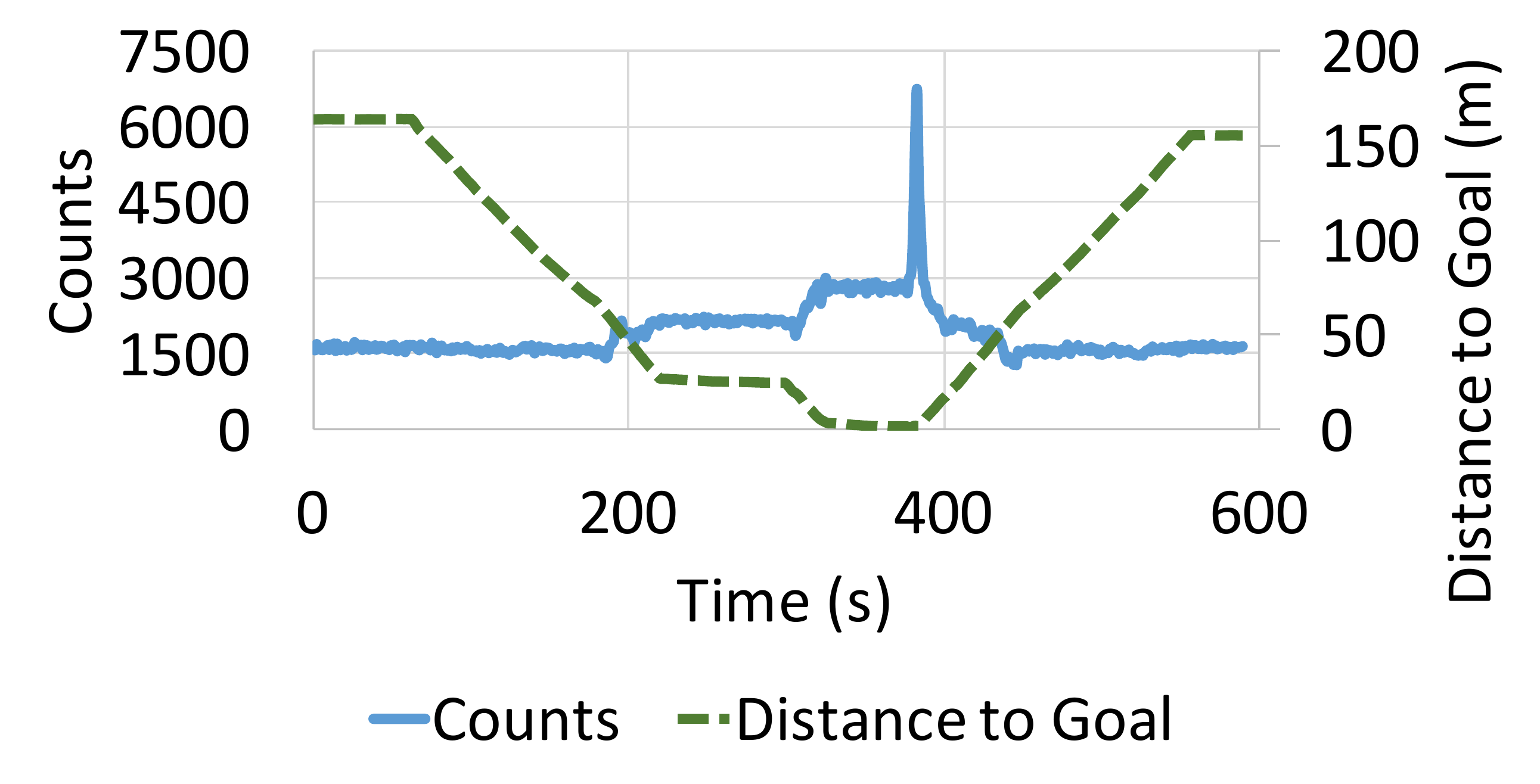}
	}
	\subfigure[Source configuration 2]{
		\label{fig:counts_config2}
		\includegraphics[width=0.48\columnwidth]{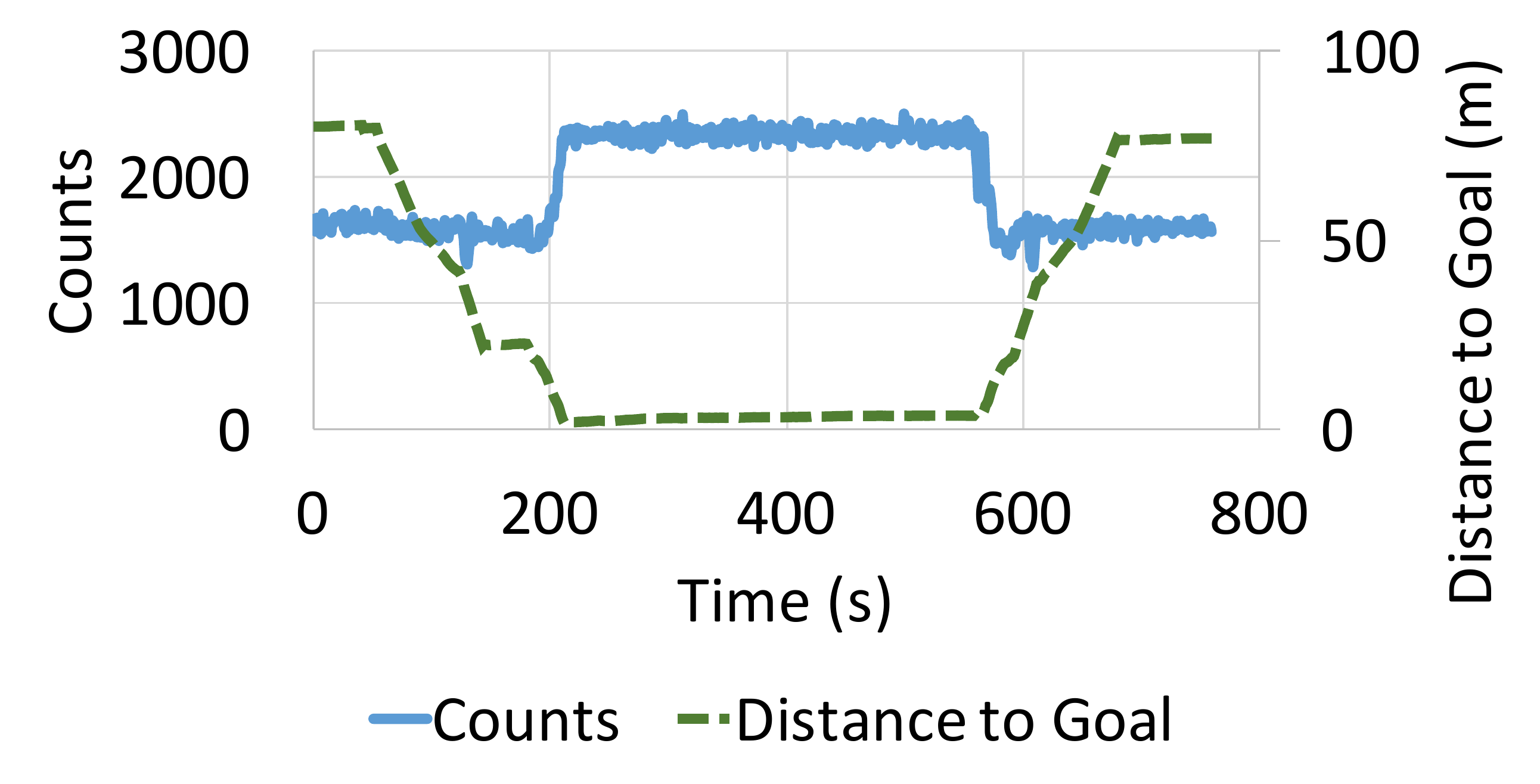}
	}
    \caption{Plots of the counts over time for both radiation source configurations. The distance to the goal position is also plotted to help understand the trends in the counts. Upon arriving at the destination, the TURTLE performed a long-dwell measurement by remaining in place for a few minutes before returning to the start position, which explains the longer period with increased counts.  The spike in (a) is believed to be a result of the TURTLE turning around to return home, during which the direction of the detector changed causing it to pick up a much stronger signal.}
    \label{fig:counts_plot}
\end{figure*}

As we approach the source in each mission, we observe a significant increase in the counts, thereby confirming that a source is present. A plot of the counts over time for each mission can be seen in \figref{fig:counts_plot}. The distance to the goal for each mission is also shown to help understand the trends of the counts, see when the TURTLE is stationary, \emph{etc}. For the counts of Mission 1, shown in \figref{fig:counts_config1}, we see a gradual increase as the TURTLE moves closer to the source before remaining in place for several minutes. The spike in the counts, observed by a single data point, was attributed to the change in the direction when the TURTLE turned around to return to the start position, resulting in a stronger signal to be seen by the radiation detector. By looking at \figref{fig:counts_config1}, the spike occurs right at the time the TURTLE ends the dwell period and begins to return to the start position.  For Mission 2, shown in \figref{fig:counts_config2}, we see a sudden increase in the counts as it reaches the destination before performing the long-dwell measurement. The source location visited in Mission 2 is not as strong as the location of Mission 1, and therefore we do not see the gradual increase seen in Mission 1.  In both cases, however, the presence of a radiation source near both locations of max counts from the aerial measurements was clearly confirmed by the TURTLE. In practical applications, images and video could be transmitted back to a remote base station where operators could take control of the TURTLE to perform additional tasks.

\section{Conclusions}
\label{sec:conclusions}

We presented an approach to the autonomous search for hazardous radiation sources in an unknown environment.
We tested our approach in a 7 acre area containing buildings, roads, grass, vegetation, \emph{etc}. To collect radiation data, elevation information, and a semantic understanding of the entire area, we used a UAV (the Yamaha RMAX) to fly over the area and collect gamma radiation data and 2D color images from off-the-shelf cameras.  The radiation data was used to output positions of the strongest reading from the detector as a destination for a UGV (the TURTLE) to visit and collect more data. The imagery was used to create a georeferenced orthophoto and DEM of the scene, which were then used to perform semantic segmentation (\ie assign a category label to each pixel in the orthophoto/DEM) with high accuracy.  By using the DEM to reason about category predictions we were able to achieve significant improvements over the 2D only baseline (orthophoto only). These image-based outputs were then used to plan a path for the TURTLE to visit the points of interest from the radiation data, where costs of the path planning algorithm were dependent on the semantic segmentation. This resulted in a preference for the TURTLE to follow roads over grass. 

After planning the paths, we deployed the TURTLE to run the two missions, where we place obstacles on each path so that it was forced to identify the obstacle and find an alternative route.  The algorithms were successfully able to identify the obstacles, update the global map, and plan a new path around the obstacles to each destination.  We also observed significant increases in the counts (sum of the 1024-d vector from the radiation detector) as the TURTLE approached the destination in each mission, confirming that sources were present in both cases. We demonstrated success for both the aerial and ground operations in our experiments to estimate and validate radiation source locations in an unknown environment. In future work, we plan to test our approach of using image-based reasoning to perform more complicated search tasks in more challenging scenes. Also, although our experiments focus on the task of autonomously searching for radiation sources, we note that this approach can be applied to many sensing tasks with the possibility of multiple aerial and ground vehicles driving the search effort.  

\postrebuttal{
We believe that with real-time 3D reconstructions from imagery, a real-time response with our system is possible. With more expensive machine vision cameras we believe we could have used existing reconstruction software to accomplish this. However, we note that we drastically reduce the price of the system with our 2 off-the-shelf Canon A810 cameras, which were triggered by an Arduino microcontroller. For future work, we are currently developing our own code to perform faster 3D reconstructions from images taken from our stereo setup with the Canon cameras by taking advantage of the known extrinsics of the imaging rig. We believe that this will help close the gap between cost and efficiency. We also believe that the annotated dataset we used to train the semantic segmentation model does generalize to many similar types of scenes, and have observed this by performing a qualitative evaluation on other test areas that we have not yet annotated to measure full performance. As more data is annotated with additional categories, and as models start to make better predictions, we believe that a system similar to the one presented in this work will become very useful for many types of disaster response scenarios.
}

\postrebuttal{
Overall the experimental results that we obtained were favorable. We did learn the importance of active search from the ground. In one of our experiments we unexpectedly failed to identify the location of the second radiation source from an altitude of 30m both by manual inspection of the counts, and with highly capable radiation detection algorithms that analyze all dimensions of the radiation data coming from the detector \cite{benedetto2014spie}. We therefore used the max counts as the estimated position of the source in each experiment. In future work we plan to expand the search from the ground to better detect these weaker sources. We still believe that our semantic maps of the area can assist in this process. For example, if we know the locations of buildings and vehicles, then the UGV can be tasked to visit these places and attempt to enter them to collect data not observed by the UAV. We may also be able to learn radiation background signatures for different semantic categories and use that to make more informed decisions about the presence or absence of a radiation source at a particular location.
}

\subsubsection*{Acknowledgments}
This work was supported by the Defense Threat Reduction Agency Basic Research Program.

\bibliographystyle{apalike}
\bibliography{gordon}

\end{document}